\documentclass[11pt]{report}
\usepackage{geometry}        
\usepackage[parfill]{parskip}  
\usepackage{graphicx}
\usepackage{amssymb}
\usepackage{epstopdf}
\usepackage{caption}
\usepackage{subcaption} 
\usepackage[toc,page]{appendix} 
\usepackage{float} 
\usepackage{datetime}

\usepackage[colorlinks=true, pdfstartview=FitV, linkcolor=blue, 
            citecolor=blue, urlcolor=blue]{hyperref}

\date{\formatdate{7}{09}{2016}}

\title{
	{Master's Thesis}\\
	{Deep Learning for Visual Recognition}
}
\author{Remi Cadene\\Supervised by Nicolas Thome and Matthieu Cord}
\begin{document}

\maketitle

\tableofcontents


\begin{abstract}
	
	The goal of our research is to develop methods advancing automatic visual recognition. In order to predict the unique or multiple labels associated to an image, we study different kind of Deep Neural Networks architectures and methods for supervised features learning. 
	We first draw up a state-of-the-art review of the Convolutional Neural Networks aiming to understand the history behind this family of statistical models, the limit of modern architectures and the novel techniques currently used to train deep CNNs.
	The originality of our work lies in our approach focusing on tasks with a low amount of data. We introduce different models and techniques to achieve the best accuracy on several kind of datasets, such as a medium dataset of food recipes (100k images) for building a web API, or a small dataset of satellite images (6,000) for the DSG online challenge that we've won.
    We also draw up the state-of-the-art in Weakly Supervised Learning, introducing different kind of CNNs able to localize regions of interest.
	Our last contribution is a framework, build on top of Torch7, for training and testing deep models on any visual recognition tasks and on datasets of any scale.

\end{abstract}

\renewcommand{\abstractname}{Acknowledgements}
\begin{abstract}
I would specifically like to thank Prof. Matthieu Cord and Assoc. Prof. Nicolas Thome for supervising my research on this project and providing resources for the experiments.\\ Additionally, I thank all the people at LIP6 for the perfect working atmosphere.\\
Lastly, I thank my family and friends for their love and support.
\end{abstract}

\chapter*{Introduction}
\addcontentsline{toc}{chapter}{Introduction}

\section*{Context}
\addcontentsline{toc}{section}{Context}

Since the beginning of the Web 2.0, the amount of visual data has grown exponentially. As an example, the director of Facebook AI Research Yann LeCun has said that almost 1 billion new photos were uploaded each day on Facebook in 2016 \footnote{\url{https://youtu.be/vlQomVlaNFg}}. Thus, computer vision has become ubiquitous in our society, with many applications such as search engine, image understanding, medicine and self-driving car. Core to many of these applications are visual recognition tasks namely image classification, localization and detection. While this seems natural to humans, those tasks are difficult due to the large number of objects in the world, the continuous set of viewpoints from which they can be viewed, the lighting in scene, color variations, background clutter, or occlusion. Sometimes those visual tasks can even be challenging for untrained humans when several classes look very similar such as in fine grained recognition, or very different such as when age, gender, version, etc. are present.

It has long been the goal of computer vision researchers to have a flexible representation of the visual world in order to recognize objects in complex scenes.
During the 2000's, the best accuracy was obtained using a hand-crafted model called the Bag of visual Words (BoW).  In a first step, robust features extractors (e.g. SIFT \cite{lowe1999object}) were applied to the dataset for extracting local descriptors from the images. Then, a clustering algorithm (e.g. K-Means) was used to obtain a visual descriptor codebook. Finally, each image were assigned to their own representation in a lower space thanks to a pooling step aggregating all the descriptors. Later, a classifier (e.g. Support Vector Machine and kernel methods) was trained on top of the vectorial representation obtained using BoW.
Recent developments in Deep Learning have greatly advanced the performance of these state-of-the-art visual recognition systems to the extent of sweeping aside the hand crafted models such as BoW. Nowadays a lot of products in the industry have benefited from the past years of research in Deep Learning. We can cite Google Photos, Flickr and Facebook, three of the world's largest photo sharing services, that use Deep Learning technologies to efficiently order and sort out piles of pictures \footnote{\url{http://googleresearch.blogspot.fr/2013/06/improving-photo-search-step-across.html}}, to better target advertising, or to find people associated to faces \cite{taigman2014deepface}. Startups also are using Deep Learning to build better recognition products and to revolutionize the market providing new services \footnote{\url{http://blog.ventureradar.com/2016/01/19/18-deep-learning-startups-you-should-know}}. Furthermore, it is used to build physical products such as self-driving cars, drones and any kind of robots equipped with cameras \footnote{\url{http://fortune.com/2015/12/21/elon-musk-interview}}.

Deep Learning can be summed up as a sub field of Machine Learning studying statical models called deep neural networks. The latter are able to learn complex and hierarchical representations from raw data, unlike hand crafted models which are made of an essential features engineering step. This scientific field has been known under a variety of names and has seen a long history of research, experiencing alternatively waves of excitement and periods of oblivion \cite{Schmidhuber2014}.
Early works on Deep Learning, or rather on Cybernetics, as it used to be called back then, have been made in 1940-1960s, and describe biologically inspired models such as the Perceptron, Adaline, or Multi Mayer Perceptron \cite{Rosenblatt1958, hebb19680}.
Then, a second wave called Connectionism came in the 1960s-1980s with the invention of backpropagation \cite{rumelhart1988learning}. This algorithm persists to the present day and is currently the algorithm of choice to optimize Deep Neural Networks.
A notable contribution is the Convolutional Neural Networks (CNNs) designed, at this time, to recognize relatively simple visual patterns, such as handwritten characters \cite{lecun1995convolutional}.
Finally, the modern era of Deep Learning has started in 2006 with the creation of more complex architectures \cite{hinton2006fast, bengio2007greedy, ranzato2007unsupervised,  Goodfellow-et-al-2016-Book}. Since a breakthrough in speech and natural language processing in 2011, and also in image classification during the scientific competition ILSVRC in 2012 \cite{Krizhevsky2012}, Deep Learning has conquered many Machine Learning communities, such as Reddit, and won challenges beyond their conventional applications area \footnote{\url{http://blog.kaggle.com/2014/04/18/winning-the-galaxy-challenge-with-convnets}}.

Especially during the last four years, Deep Learning has made a tremendous impact in computer vision reaching previously unattainable performance on many tasks such as image classification, objects detection, object localization, object tracking, pose estimation, image segmentation or image captioning \cite{DBLP:journals/corr/GuWKMSSLWW15}.
This progress have been made possible by the increase in computational resources, thanks to frameworks such as Torch7, modern GPUs implementations such as Cudnn, the increase in available annotated data, and the community-based involvement to open source codes and to share models. These facts allowed for a much larger audience to acquire the expertise needed to train modern convolutional networks. Thus, larger and deeper architectures are trained on bigger datasets to achieve better accuracy each year. Also, already trained models have shown astonishing results when transfered on smaller datasets and evaluated on different visual tasks. Hence, a lot of pretrained models are available on the web. 
However, CNNs still possess inherent limitations. From a theoretical perspective, Deep Neural Networks are not well understood due to their non convex property. Despite numerous efforts, a proof of convergence to good local minima has never been found. Thus, most of the research made in this field are experimentally driven and empirical \cite{zeiler2014visualizing}.
From a practical perspective, their need for large amounts of training samples does not provide them the ability to generalize when trained on small and medium datasets. In this context, weakly supervised learning methods, that we describe in the next chapter, can be applied to overcome this limitations.
Nevertheless, Deep Neural Networks seems to be the most promising kind of models for solving visual recognition.

The progress, needs and expectations of Deep Learning are undoubtedly signs of the Big Data era, where images of any kind and computable capabilities are more available than ever before. In this context, the Convolutional Neural Networks are the most efficient statistical model for visual recognition. The goal of this work is to produce an analysis of the state-of-the-art methods to train such models, to explain their limitations and to propose original idea to overcome the latter.

\section*{Contributions}
\addcontentsline{toc}{section}{Contributions}

This master's thesis introduces a number of contributions to different aspects of visual recognition. However our work is focused on classifying images and recognizing objects using global labels (e.g. one label to indicates the presence or absence of the object).

\begin{itemize}
	\item In the first chapter, we draw up the state of the art explaining how Convolutional Neural Networks (CNNs) achieve such good accuracy, describing different architectures and clarifying their limits. In the last chapter, we also draw up the state of the art in Weakly Supervised Learning which gather methods to improve the accuracy of CNNs trained on a few amount of images.
	
	\item In the second chapter, we apply Deep Neural Networks on a medium dataset of food recipes. We notably show that training CNNs From Scratch with large amount of parameters is possible on this kind of datasets reaching way better accuracy than the BoW model. We also show that Fine Tuning of pretrain models is essential, especially on small dataset. We illustrate this last point by presenting our winning solution of the Data Science Game Online Selection, an image classification challenge made for master and PhD students from all around the world.
	
	\item In the last chapter, we study techniques to overcome the limited spatial invariance capacity of CNNs without the use of rich annotations such as bounding boxes. The first technique consists in providing such invariance directly by feeding the networks augmented images. The second consists in using a novel approach developed by a PhD student at LIP6 which gives the network the ability to localize regions of interest. The third one consists in using a first network to localize precisely the object and a second network to classify the resulting image. Our main results are synthesized at the end of this chapter.
	
	\item Overall, this study has helped us to develop \textit{Torchnet-vision}, a framework build on top of Torch7 that serves as a plugin for Torchnet (a high level deep learning framework) for training easily the last architectures and pretrain models on several datasets. Reproducibility of a large amount of our experiments is ensured by the fact that we provide links to our code in footnotes.
\end{itemize}

\chapter{Convolutional Neural Networks}




%


\section{Layers}

\subsection{Linear or Fully Connected}

Mathematically, we can think of a linear layer as a function which applies a linear transformation on a vectorial input of dimension $I$ and output a vector of dimension $O$. Usually the layer has a bias parameter.
$$y = A \bullet x + b$$
$$y_i = \sum^{I}_{j=1} (A_{i,j}  x_j) + b_i$$

The linear layer is motivated by the basic computational unit of the brain called neuron. Approximately 86 billion neurons can be found in the human nervous system and they are connected with approximately $10^{14}$ - $10^{15}$ synapses. Each neuron receives input signals from its dendrites and produces output signal along its axon. The linear layer is a simplification of a group of neuron having their dendrites connected to the same inputs. Usually an activation function, such as sigmoid, is used to mimic the 1-0 impulse carried away from the cell body and also to add non linearity. However we consider here that the activation function is the identity function that output real values. 

\begin{figure}
	\begin{subfigure}{.50\textwidth}
		\centering
		\includegraphics[width=.95\linewidth]{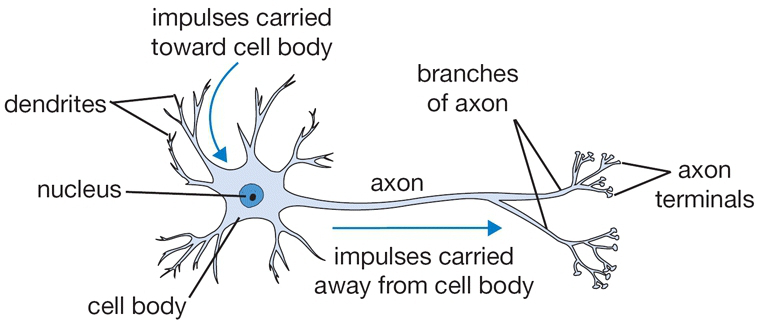}
		\caption{}
		\label{fig:sfig1}
	\end{subfigure}%
	\begin{subfigure}{.50\textwidth}
		\centering
		\includegraphics[width=.95\linewidth]{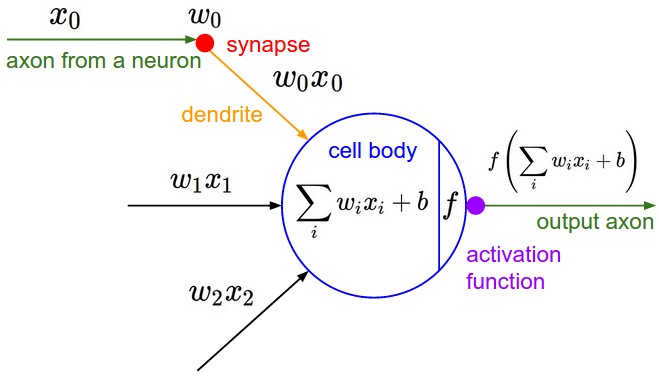}
		\caption{}
		\label{fig:sfig2}
	\end{subfigure}
	\caption{A cartoon drawing of a biological neuron (a) and its mathematical model (b).}
	\label{fig:2images}
\end{figure}

\subsection{Activation or Non Linearity}

The capacity of the neural networks to approximate any functions, especially non-convex, is directly the result of the non-linear activation functions. Every kind of activation function takes a vector and performs a certain fixed point-wise operation on it. There are three main activation functions.

\paragraph{Sigmoid}
The Sigmoid non-linearity has the following mathematical form $$y = \sigma(x) = 1/(1 + \exp^{-x})$$
It takes a real value and squashes it between 0 and 1. However, when the neuron's activation saturates at either tail of 0 or 1, the gradient at these regions is almost zero. Thus, the backpropagation algorithm fail at modifying its parameters and the parameters of the preceding neural layers.
	
\paragraph{Hyperbolic Tangent} The TanH non-linearity has the following mathematical form $$y = 2 \sigma(2 x) - 1$$ It squashes a real-valued number between -1 and 1. However it has the same drawback than the sigmoid.

\paragraph{Rectified Linear Unit} The ReLU has the following mathematical form $$y = max(0,x)$$ The ReLU has become very popular in the last few years, because it was found to greatly accelerate the convergence of stochastic gradient descent compared to the sigmoid/tanh functions due to its linear non-saturating form (e.g. a factor of 6 in \cite{Krizhevsky2012}).
	In fact, it does not suffer from the vanishing or exploding gradient. An other advantage is that it involves cheap operations compared to the expensive exponentials. However, the ReLU removes all the negative informations and thus appears not suited for all datasets and architectures.

\begin{figure}
	\begin{subfigure}{.45\textwidth}
		\centering
		\includegraphics[width=.65\linewidth]{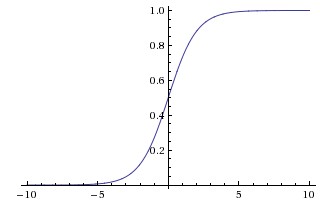}
		\caption{}
		\label{fig:sfig1}
	\end{subfigure}%
	\begin{subfigure}{.45\textwidth}
		\centering
		\includegraphics[width=.65\linewidth]{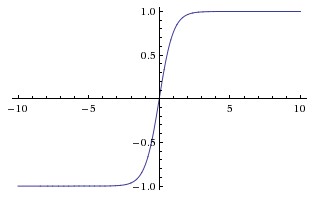}
		\caption{}
		\label{fig:sfig2}
	\end{subfigure}
	\caption{A sigmoid (a) and a tanh (b).}
	\label{fig:2images}
\end{figure}

\subsection{Spatial Convolution}

Regular Neural Networks, only made of linear and activation layers, do not scale well to full images. For instance, images of size $3 \times 224 \times 224$ (3 color channels, 224 wide, 224 high) would necessitate a first linear layer having $3*224*224 + 1 = 150,129$ parameters for a single neurone (e.g. output). Spatial convolution layers take advantage of the fact that their input (e.g. images or feature maps) exhibits many spatial relationships. In fact, neighboring pixels should not be affected by their location within image. Thus, a convolutional layer learns a set of $N_k$ filters $F = {f_1,...,f_{N_k}}$, which are convolved spatially with input image $x$, to produce a set of $N_k$ 2D features maps $z$:
$$z_k = f_k \ast x$$
where $\ast$ is the convolution operator. When the filter correlates well with a region of the input image, the response in the corresponding feature map location is strong. Unlike conventional linear layer, weights are shared over the entire image reducing the number of parameters per response and equivariance is learned (i.e. an object shifted in the input image will simply shift the corresponding responses in a similar way).
Also, a fully connected layer can be seen as a convolutional layer with filter of sizes $1\times1\times inputSize$.

It is important to highlight that a spatial convolution is not defined by the spatial size of the input feature maps (e.g. wide and high), neither by the size of the output feature maps, but by the number of filters (e.g. number of output channels),  the properties of its filters (e.g. number of input channels, wide, high) and the properties of the convolution (e.g. padding, stride). Animations showing different kind of convolution can be viewed on line \footnote{\url{https://github.com/vdumoulin/conv_arithmetic}}.

\subsection{Spatial Pooling}

In Convolutional Neural Networks, a pooling layer is typically present to provide invariance to slightly different input images and to reduce the dimension of the feature maps (e.g. wide, high):
$$p_R = P_{i \in R} (z_i)$$
where $P$ is a pooling function over the region of pixels $R$. Max pooling is preferred as it avoids cancellation of negative elements and prevents blurring of the activations and gradients throughout the network since the gradient is placed in a single location during backpropagation.

The spatial pooling layer is defined by its aggregation function, the high and width dimensions of the area where it is applied, and the properties of the convolution (e.g. padding, stride).

\begin{figure}
	\centering
	\includegraphics[width=.35\linewidth]{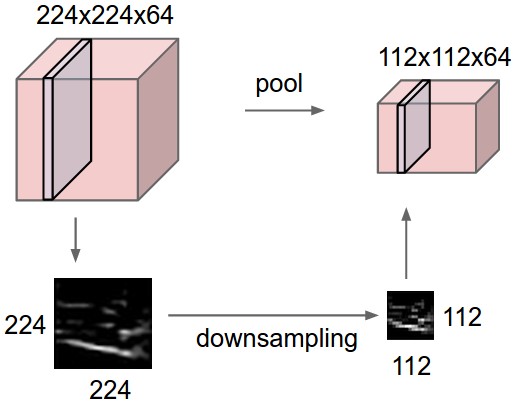}
	\caption{The illustration of a spatial pooling operation in $2\times2$ regions by a stride of 2 in the high direction, and 2 in the width direction, without padding.}
	\label{fig:2images}
\end{figure}



\subsection{Batch Normalization}

This layer quickly became very popular mostly because it helps to converge faster \cite{DBLP:journals/corr/IoffeS15}. It adds a normalization step (shifting inputs to zero-mean and unit variance) to make the inputs of each trainable layers comparable across features. By doing this it ensures a high learning rate while keeping the network learning.

Also it allows activations functions such as TanH and Sigmoid to not get stuck in the saturation mode (e.g. gradient equal to 0).

\section{Convolutional Architectures}

A lot of convolutional architectures have been developed from the 1990's. In this section, we make an inventory of the most known architectures \footnote{\url{http://culurciello.github.io/tech/2016/06/04/nets.html}}. Each one represent a step further for more advanced visual recognition.

\subsection{CNNs (LeNet)}

\paragraph{LeNet-5}
This kind of architecture is one of the first successful applications of CNNs. It was developed by Yann LeCun in the 1990's and was used to read zip codes and digits. This architecture, with regard to the modern ones, differs on many points. Thus, we will limit ourselves on the most known, LeNet-5 \cite{lecun1998gradient}, and we will not delve into the details. In overall this network was the origin of much of the recent architectures, and a true inspiration for many people in the field.

LeNet-5 features can be summarized as:
\begin{itemize}
	\item sequence of 3 layers: convolution, pooling, non-linearity,
	\item inputs are normalized using mean and standard deviation to accelerate training  \cite{le1991eigenvalues},
	\item sparse connection matrix between layers to avoid large computational cost
	\item hyperbolic tangent or sigmoid as non-linearity function,
	\item trainable average pooling as pooling function,
	\item fully connected layers as final classifier,
	\item mean squared error as loss function.
\end{itemize}


\begin{figure}[h]
	\centering
	\includegraphics[width=.95\linewidth]{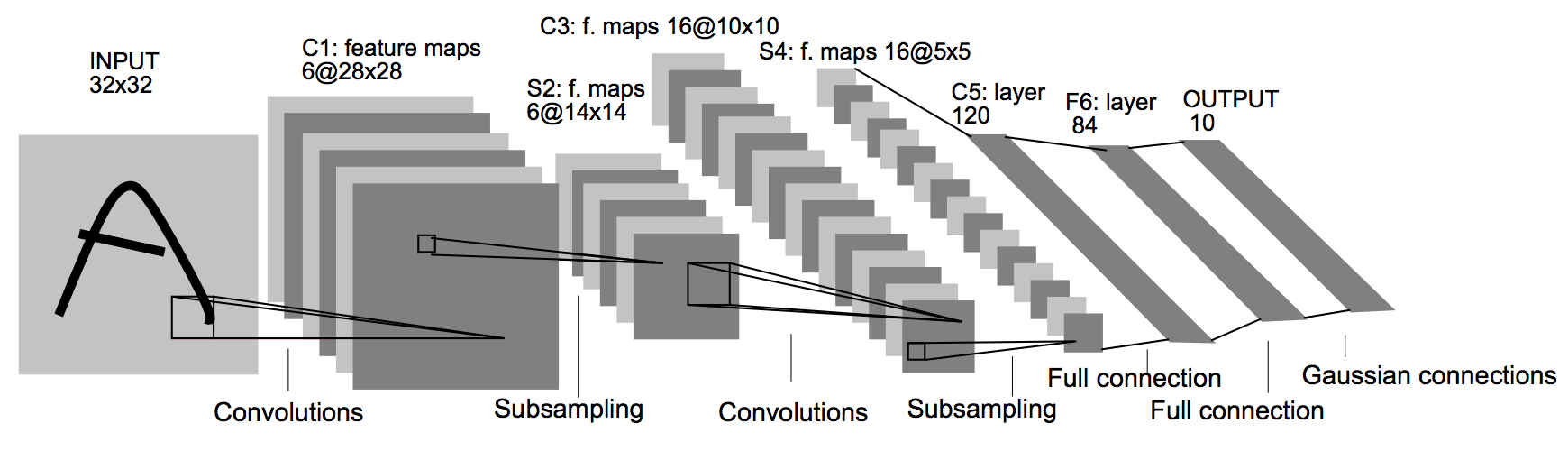}
	\caption{Architecture of LeNet-5, an old convolutional neural network for digits recognition. }
	\label{fig:2images}
\end{figure}

\subsection{Deep CNNs}

\paragraph{AlexNet}
It is one of the first work that popularized convolutional networks in computer vision. AlexNet \cite{krizhevsky2012imagenet} was submitted to the ImageNet ILSVRC challenge of 2012 and significantly outperformed the other hand crafted models (accuracy top5 of 84\% compared to the second runner-up with 74\%). This network, compared to LeNet, was deeper (60 millions of parameters) and bigger (5 convolutional layers, 3 max pooling and 3 fully-connected layers). At this time, the authors provided a multi-GPUs implementation in CUDA to bypass the memory needs.
It popularized:
\begin{itemize}
	\item the ReLU as non-linearity function of choice,
	\item the method of stacking convolutional layers plus non-linearity on top of each other without being immediately followed by a pooling layer,
	\item  the method of overlapping Max Pooling, avoiding the averaging effects of Average Pooling.
\end{itemize}

\begin{figure}[h]
	\centering
	\includegraphics[width=.95\linewidth]{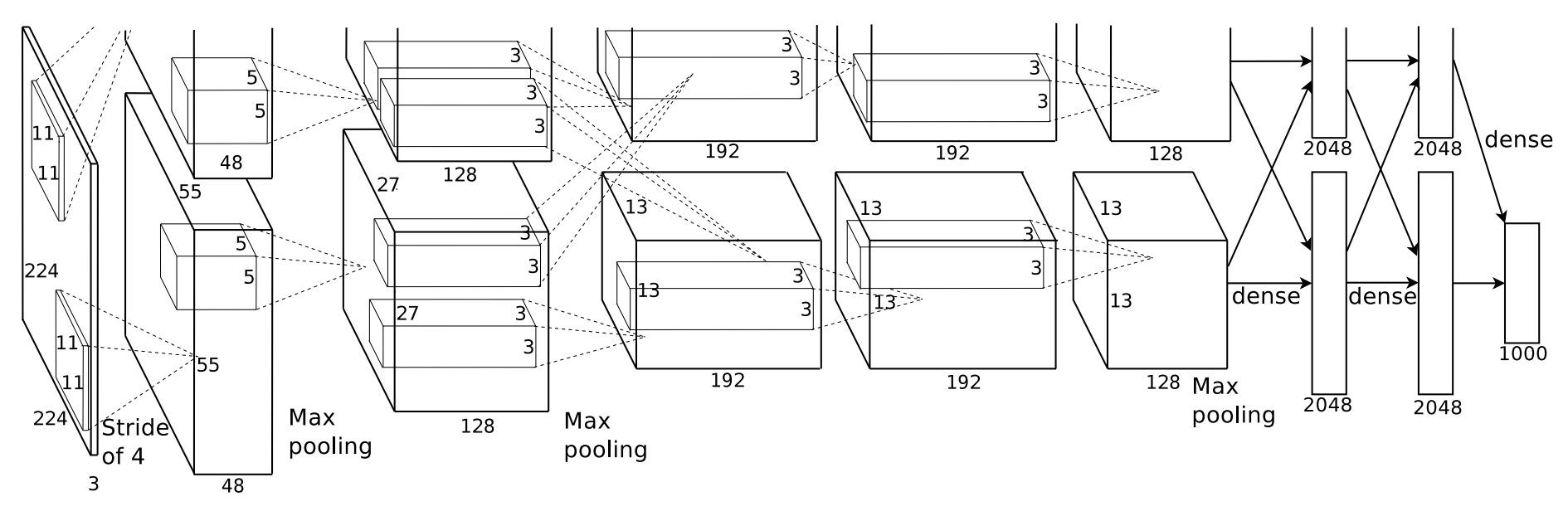}
	\caption{An illustration of the architecture of AlexNet. One GPU runs the layer-parts at the top of the figure while the others runs the layer-parts at the bottom.. }
	\label{fig:2images}
\end{figure}

\paragraph{Overfeat or ZFNet} It was the winner architecture of ILSVRC2013 \cite{sermanet2013overfeat} with almost 140 millions of parameters. Based on AlexNet, the size of its middle convolutional layers have been expanded. Also, the stride and filter size on its first layer have been made smaller.

\subsection{Very Deep CNNs}

\paragraph{VeryDeep or VggNet} It was the runner-up architecture of ILSVRC2014 \cite{simonyan2014very} with almost 140 millions of parameters. Its main contributions were to show that depth is a critical component for good performance, to use much smaller $3\times3$ filters in each convolutional layers, and also to combine them as a sequence of convolutions. The great advantage of VggNet was the insight that multiple $3\times3$ convolution in sequence can emulate the effect of larger receptive fields, for examples $5\times5$ and $7\times7$. These ideas will be also used in more recent network architectures as Inception and ResNet.

\begin{figure}[h]
	\centering
	\includegraphics[width=.30\linewidth]{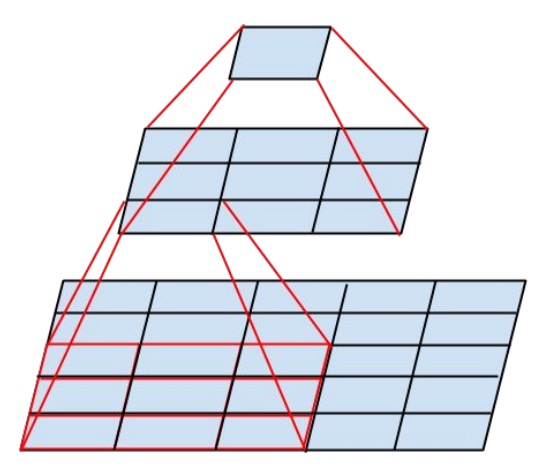}
	\caption{Filter of $5 \times 5$ or more can be decomposed with multiple $3 \times 3$ convolutions such as in VGG.}
	\label{fig:2images}
\end{figure}

\paragraph{GoogLeNet or Inception} It was the winner architecture of ILSVRC2014 \cite{szegedy2015going}. Its main contribution was the development of an Inception Module that dramatically reduced the number of parameters (40 millions) \cite{url-onebyoneconv}. Also, it eliminated a large amount of parameters by using average pooling instead of fully connected layers at the top of the convolutional layers. Further versions of the GoogLeNet has been released. The most recent architecture available is InceptionV3 \cite{szegedy2015rethinking}. Notably, it uses batch normalization.

\begin{figure}[h]
	\centering
	\includegraphics[width=1\linewidth]{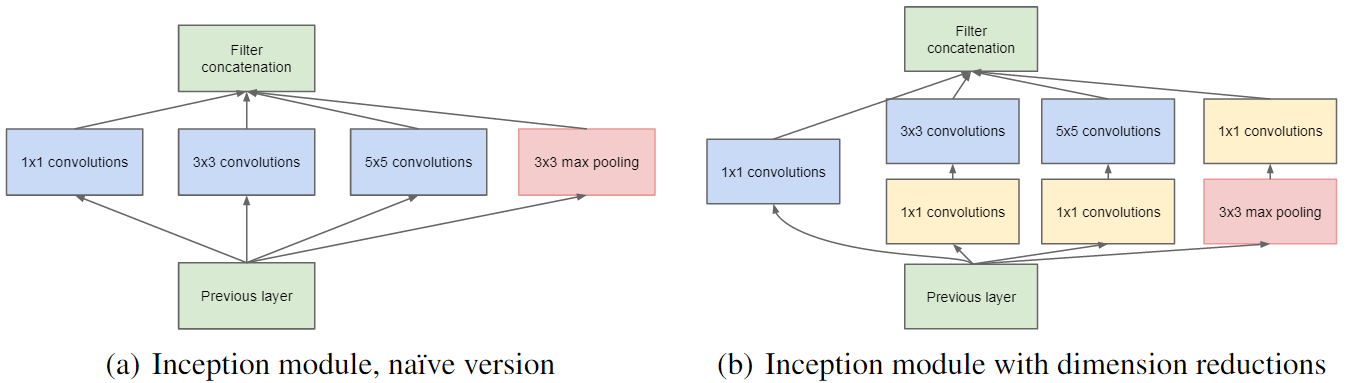}
	\caption{$1 \times 1$ convolutions are used to decrease the input size before $3 \times 3$ convolutions in order to provide more combinational power such as in GoogLeNet.}
	\label{fig:2images}
\end{figure}

\subsection{Residual CNNs}

\paragraph{ResNet} It was the winner architecture of ILSVRC2015 \cite{Ioffe2015} with 152 layers. Its main contribution was to use batch normalization and special skip connections for training deeper architectures. ResNet with 1000 layers can be trained with those techniques. However, it has been empirically found that ResNet usually operates on blocks of relatively low depth ($\sim$20 - 30 layers), which act in parallel, rather than serially flow the entire length of the network \cite{DBLP:journals/corr/VeitWB16}.

\begin{figure}
	\centering
	\includegraphics[width=.57\linewidth]{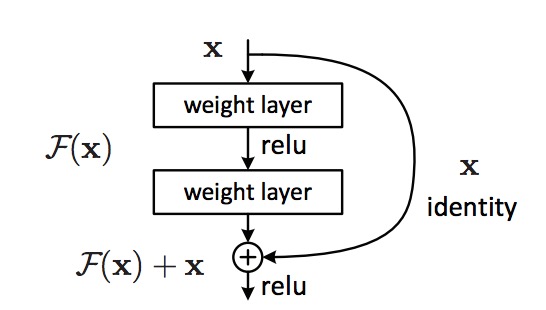}
	\caption{A skip connection is used to bypass the input to the next layers such as in ResNet.}
	\label{fig:2images}
\end{figure}

\section{Training Methods}

\subsection{From Scratch}

\paragraph{Initialization}

All the network parameters are generally initialized with Layer-sequential unit-variance (LSUV) (e.g. each parameters as Gaussian random variables with mean 0 and standard deviation $\frac{1}{n_{inputs}}$ and biases are initialized to zero). Since the LSUV initialization works under assumption of preserving unit variance of the input, pixel intensities are given after subtracting the mean and dividing by the standard deviation. More information can be found in the chapter 3 of Michael Nielsen's book \cite{url-booknielsen}.
In case of pretrain networks, the mean and std of the original dataset are kept.

\paragraph{Loss function} To quantify the capacity of the network to approximate the ground truth labels for all training inputs, we define a loss function which takes as inputs the weights, biases, and examples from the training set. For instance, the loss could be the number of images correctly classified. However, the most efficient way to find the weights and biases, regarding the number of parameters, is to use an algorithm similar to the Stochastic Gradient Descent (SGD). In order to do so, if our chosen loss function is not smooth, we have to chose a surrogate loss (e.g. derivable function) such as Mean Square Error or Cross Entropy.

\paragraph{Backpropagation}
For each examples, we compute the prediction and its associated loss. We sum up all the loss to compute the final error. Then we use the backpropagation algorithm to propagate the error in order to compute the partial derivatives $\frac{\delta E}{\delta w} $ and $\frac{\delta E}{\delta b}$ of the cost function $E$ for all weights $w$ and bias $b$. In this work, our goal is not to explain in details how works the backpropagation algorithm. We advise the curious reader to read the chapter 2 of Michael Nielsen's book \cite{url-booknielsen}.

\paragraph{Optimization}
Once all the derivatives are computed, we update our parameters using a chosen optimization technique such as SGD.
We then iterate the predication (e.g. forward pass), the backpropagation of errors (e.g. backward pass) and the optimization until convergence hopping to find a local minimum low enough to ensure good predictions. Even if the chosen surrogate loss function of a neural network is non convex, SGD works well in practice.

\paragraph{Grid search}
It is common to explore manually the space of hyperparameters such as learning rate, weight decay, learning rate decay, amount of dropout, not to mention the architectures hyperparameters, in order to obtain the best performance in terms of both accuracy and training time.

\cite{mishkin2016systematic} made an evaluation on the influence of architecture choices and optimization hyperparameters on ImageNet. While there are very few theoretical studies, technical studies of this kind can help the reader to reduce the space of hyperparameters to explore.

\subsection{Transfer Learning}

\paragraph{Features Extraction}
It consists in extracting features from the network by forwarding examples. Transformations to the examples are possible such as horizontal flip. Then the associated features to the example are aggregated whether by averaging or stacking them. Finally, a classifier is trained and tested on the features. Typically, the later is a Support Vector Machine with a linear kernel.

\paragraph{Fine Tuning}
It consists in training a pretrained network on a smaller dataset. Typically, the last fully connected layers, which can be viewed as classification layers, are reset and a smaller learning rate is applied to the pretrained layers. By doing so, the goal is to adapt the features to the new dataset. More different is the latter from the original dataset, more parameters/layers must be reset. 

\subsection{Loss functions}

In this subsection, we present the three most used loss function to train deep neural networks for classification.

\paragraph{Mean Square Error (MSE)} It is a multi class loss formerly used to train neural networks.
$$
Loss(x, y) = \frac{1}{n} \sum_i |x_i - y_i|^2
$$
with $x$ a vector of $n$ predictions, and $y$ a binary vector full of $0$ besides a $1$ in the corresponding class dimension .

\paragraph{Cross Entropy} It is a multi class loss which is nearly a better choice than MSE. $$
Loss(x, y) = - \sum_i y_i * log( \frac{exp(x_i)}{(\sum_j exp(x_j))})
$$
with $x$ a vector of $n$ predictions, and $y$ a binary vector full of $0$ besides a $1$ in the corresponding class dimension .

In fact, it may happen that the initialization of the parameters result in the network being decisively wrong for some training input (an output neuron will have saturated near 1, when it should be 0, or vice versa). The MSE loss will usually slow down learning, but the Cross Entropy loss won't. More information can be found in the chapter 3 of \cite{url-booknielsen}.

\paragraph{Loss Multi Label} It is the adaptation of the Cross Entropy loss for multi-label classification. It is a multi-label one-versus-all loss based on max-entropy.
$$
Loss(x, y) = - \sum_i (y_i * log( \frac{exp(x_i)}{1 + exp(x_i)}) + (1 - y_i) * log(\frac{1}{1+exp(x_i)}))
$$ 

\subsection{Optimization algorithms}

The loss function of a CNN is highly non convex. Hopefully the latter is also fully derivable, so that gradient based optimization algorithms can be applied. However, CNNs are usually made of tens of millions of parameters. Thus, only the first order derivatives are used in practice. In fact, the second derivatives are costly in term of memory and computational effort.

\paragraph{Stochastic Gradient Descent (SGD)} It is the main optimization algorithm. It consists in using a few examples to compute the gradient of the parameters with respect to the loss function :
$$
\theta_ {t+1} = \theta_{t} - \lambda \cdot \nabla_{\theta_t} L(f_{\theta_t}(x_i), y_i)
$$
There is no proof of good convergence. However, this algorithm reaches good local minima in practice, even when the parameters are randomly initialized. One of the reason could be the stochastic property of this algorithm, allowing the latter to optimize different loss functions and thus to get out of bad minima. The other reason could be that a lot of local minima are almost as accurate than the global minima. Answers to this question are still under active research.

\paragraph{Approximation of Second Order Derivatives} Other optimization algorithms rely on more advance techniques such as momentum, second order approximation and adaptive learning rates \cite{sutskever2013importance, DBLP:journals/corr/KingmaB14}. They are known to converge faster and their parameters are sometimes easier to tune by grid search. However, they take a bit more processing time to compute, but also much more memory (2 to 3 more).

\paragraph{Distributed SGD} It is the kind of optimization used in parallel computing environments. Different computers train the same architecture with almost the same parameters values. It allows more exploration of the parameters space, which can lead to improved performance \cite{DBLP:journals/corr/ZhangCL14a}.

\subsection{Regularization Approaches}

Deep and large enough neural networks can memorize any data. During training, their accuracy on the trainset typically converges towards perfection while it degrades on the testset. This phenomenon is called overfitting.

\paragraph{Regularization L2} The first main approach to overcome overfitting is the classical weight decay, which adds a term to the cost function to penalize the parameters in each dimension, preventing the network from exactly modeling the training data and therefore help generalize to new examples:
$$
Err(x, y) = Loss(x, y) + \sum_i \theta_i^2
$$ 
with $\theta$ a vector containing all the network parameters.
	
\paragraph{Data augmentation} It is a method of boosting the size of the training set so that the model cannot memorize all of it. This can take several forms depending of the dataset. For instance, if the objects are supposed to be invariant to rotation such as galaxies or planktons, it is well suited to apply different kind of rotations to the original images.
	
\paragraph{Dropout} Finally, a recent success has been shown with a regularization technique called Dropout \cite{srivastava2014dropout}. The idea is to randomly set a certain percentage of the activations in each layers to 0. During the training, neurons must learn better representations without co-adapting to each other being active. During the testing, all the neurons are used to compute the prediction and Dropout acts like a form of model averaging over all possible instantiations of the model.

\paragraph{Early stopping} It consists in stopping the training before the model begins to overfit the training set. In practice, it is used a lot during the training of neural networks.





\section{Interpretability}

\subsection{Definitions}

The desire for interpretation presupposes that predictions alone do not suffice \cite{lipton2016mythos}.
Typically, we train models to achieve strong predictive power. However, this objective can be a weak surrogate for the real-world goals of machine learning practitioners.

\paragraph{Intelligibility}
There are two definitions of interpretability. The first one is linked with understandability or intelligibility, i.e., that we can grasp how the model works. Multiple criteria are used to evaluate if a model is interpretable or not. For instance: \textit{will it converge? do we understand what each parameters represents? is it simple enough to be examined all at once by a human?} Understandable models are sometimes called transparent, while incomprehensible models are called black boxes.

\paragraph{Post-hoc interpretability} Discussions of interpretability sometimes suggest that human decision-makers, despite being black boxes, are themselves interpretable because they can explain their actions. Deep learning models are also often considered as black boxes. However visualization techniques can help to generate post-hoc interpretations in order to explain their actions.

\subsection{Simple Visualization Techniques}

 The first technique consists in visualizing the images directly. For instance, computing the loss over all the testing set allows to visualize the easiest or hardest examples for the network. Also, computing the activations of a certain layer to identify the k-nearest neighbors based on the proximity in the space learned by the model can be a good way to understand what the chosen layer has learned.

A similar approach consists to visualize high-dimensional distributed representations with t-SNE \cite{maaten2008visualizing}, a technique that renders 2D visualizations in which nearby data points are likely to appear close together.

\begin{figure}[h]
	\centering
	\includegraphics[width=.57\linewidth]{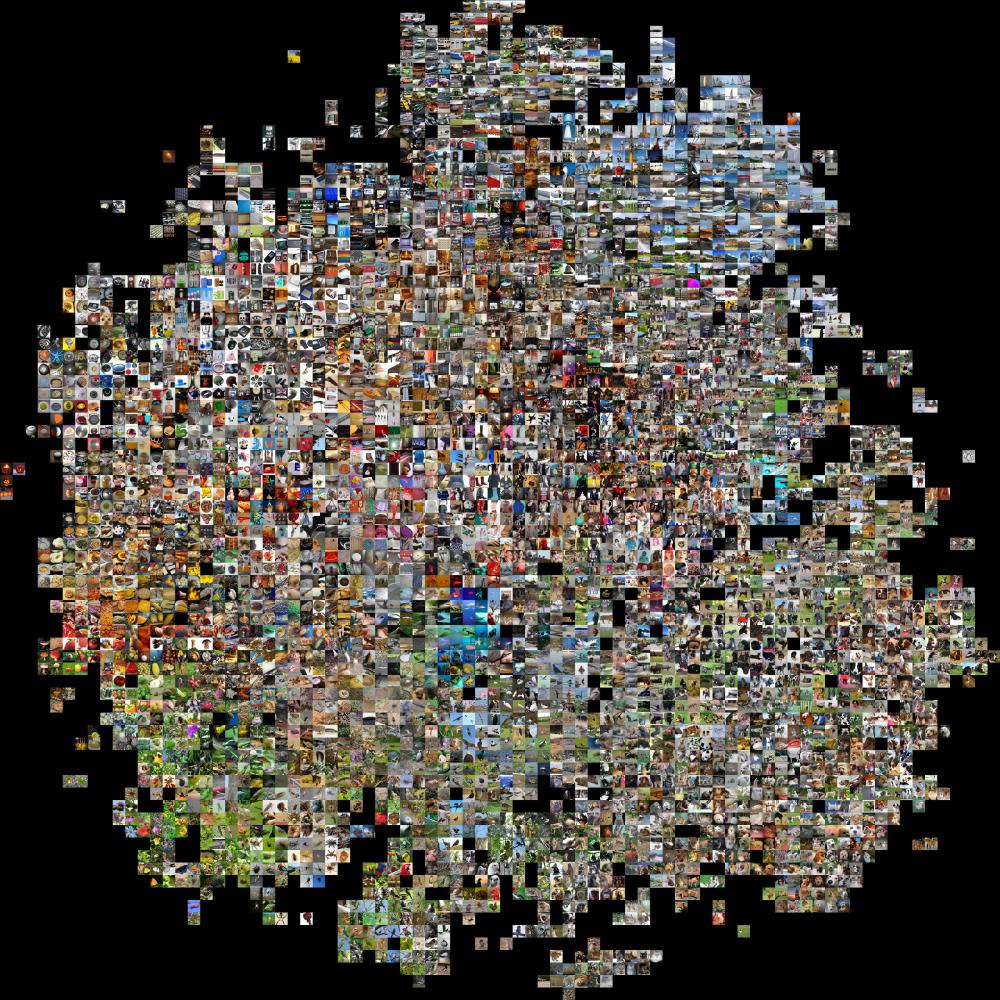}
	\caption{Embedded images from ImageNet in 2D space using t-SNE and features extracted from a CNN.}
	\label{fig:2images}
\end{figure}

An other technique consists in visualizing the features map generated by the network. However, this method is not suited when the network is large. 

\subsection{Advanced Visualization Techniques}

\paragraph{Gradient based} A popular approach is to render the gradient as an image \cite{simonyan2013deep}. While this does not say precisely how a model works, it conveys which image regions the current classification depends upon most heavily.

\begin{figure}
	\centering
	\includegraphics[width=.57\linewidth]{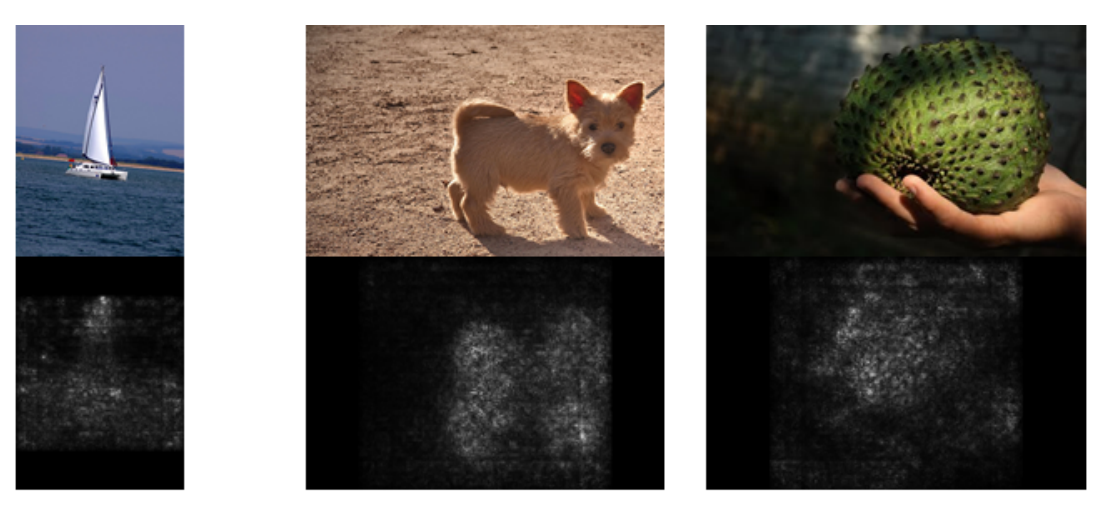}
	\caption{Original images with their associated gradient based images.}
	\label{fig:2images}
\end{figure}

\cite{mordvintsev2015inceptionism} attempt to explain what a network has learned by altering the input through gradient descent to enhance the activations of certain nodes selected from the hidden layers. An inspection of the perturbed inputs can give clues to what the models has learned.

An other technique consists in training deconvolutional neural networks \cite{zeiler2014visualizing, zeiler2010deconvolutional}. One example of using this powerful technique is to understand what CNNs are looking at when they see nudity \footnote{\url{http://blog.clarifai.com/what-convolutional-neural-networks-see-at-when-they-see-nudity}}.

\begin{figure}
	\centering
	\includegraphics[width=.57\linewidth]{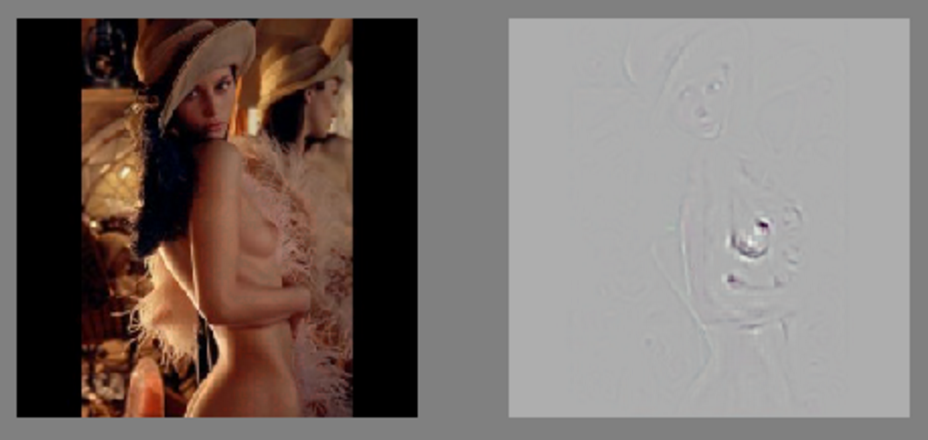}
	\caption{The original Lena photo and the resulting deconvolutional image which added to the original could improved Lena to look more like pornography (The goal is to understand what a CNN has learned). }
	\label{fig:2images}
\end{figure}

\paragraph{Semantic parts localization}
Finally other approaches study whether CNNs learn semantic parts of object classes in their internal representation. They investigate the responses of convolutional filters and try to associate their stimuli with semantic parts.

A recent study \cite{gonzalez2016semantic} discovered that the discriminative power of the network can be attributed to a few discriminative filters specialized to each object class. Despite promoting the emergence of filters learn to respond to semantic parts (and not only object class), they found that only 34 out of 123 semantic parts in PASCAL-Part dataset emerge in AlexNet. Also networks trained for image classification produced the same results than those trained for objects detection or localization.

\begin{figure}
	\centering
	\includegraphics[width=.57\linewidth]{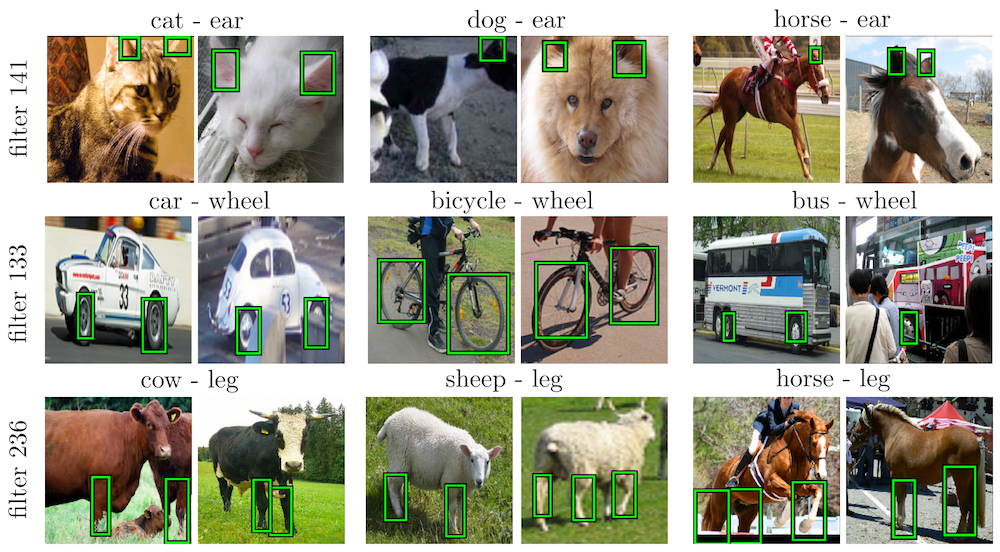}
	\caption{Detections performed by filters of AlexNet. The filters are specific to a part and they work well on several object classes containing it.}
	\label{fig:2images}
\end{figure}

\section{Invariance}

Convolutional Neural Network is a powerful model able to learn the needed invariance directly from the data. However, when the amount of labeled images (e.g. training set) is not big enough, the architectural choices can limit the capacity of the model to generalize to unknown images (e.g. testing set), or rather the opposite, to improve both the accuracy and training time.

In this section, we explain how CNNs achieve translation, rotation and scale invariance, and we propose some methods to achieve better invariance.
Note that the pooling regimes make convolution slightly invariant to translation, rotation and shifting. However, it is not sufficient to be invariant to large spatial transformations.

\subsection{Translation invariance}

Pooling, combined with striding, is a common way to achieve a degree of invariance, but even without this technique, the convolution filters and the fully-connected layers are able to learn spatial invariance \cite{url-quorarolt}. We explain our thinking in the next paragraphs.

\begin{figure}[h]
	\centering
	\includegraphics[width=.80\linewidth]{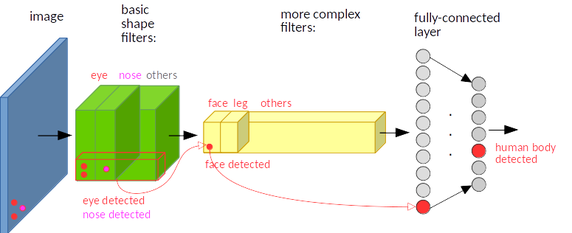}
	\caption{Ideal CNN with a human face at the bottom left of the image. }
	\label{fig:quora-invariance-1}
\end{figure}

\begin{figure}[h]
	\centering
	\includegraphics[width=.80\linewidth]{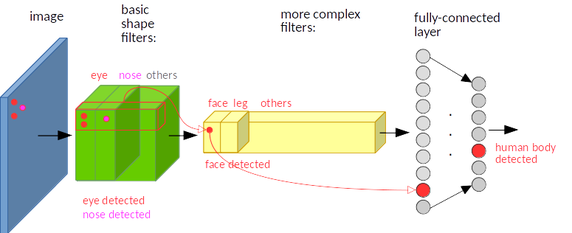}
	\caption{Ideal CNN with a human face at the top left of the image. }
	\label{fig:quora-invariance-2}
\end{figure}

In figures \ref{fig:quora-invariance-1} and \ref{fig:quora-invariance-2}, we can see an ideal network with two convolutional layers with pooling, and two fully-connected layers.
\begin{itemize}
	\item The first layer filters (which generate the green volume) detect eyes, noses and other basic shapes (in real CNNs, first layer filters match lines and very basic textures).
	\item The second layer filters (which generate the yellow volume) detect faces, legs and other objects that are aggregations of the first layer filters (real life convolution filters may detect objects that have no meaning to humans).
\end{itemize} 

In figure \ref{fig:quora-invariance-1}, a face is at the corner bottom left of the image (represented by two red and a magenta point). In figure \ref{fig:quora-invariance-2}, the same face is at the corner top left of the image. The same number of activations occurs, but they occur in different regions of the green and yellow volumes. Therefore, any activation point at the first slice of the yellow volume means that a face was detected, independently of the face location. Then the fully-connected (FC) layer is responsible to "translate" a face and two arms to an human body. In each examples, the activation path inside the FC layer was different, meaning that a correct learning at the FC layer is essential to ensure the spatial invariance property.

In conclusion, if a CNN is trained showing faces only at one corner, during the learning process, the fully-connected layer may become insensitive to some faces in other corners. Thus, regarding the task and the dataset, it can be suitable to use data augmentation to provide the CNN more examples of the same objects in different image regions. However, for small datasets in which objects can undergo strong translations in the image, some different approaches are more suitable \cite{durand2016weldon, DBLP:journals/corr/BilenV15}.

\subsection{Rotation invariance}

\begin{figure}[h]
	\centering
	\includegraphics[width=.55\linewidth]{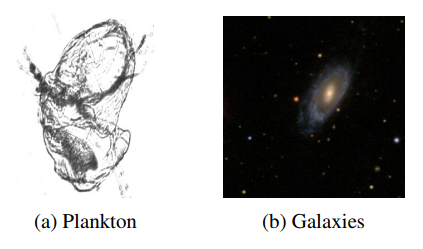}
	\caption{Example images for the Plankton and Galaxies datasets, which are rotation invariant.}
	\label{fig:rotation-invariance-1}
\end{figure}

\begin{figure}[h]
	\centering
	\includegraphics[width=.55\linewidth]{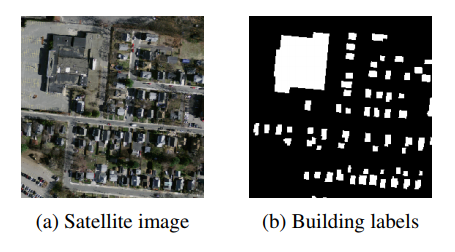}
	\caption{Example tile from the Massachusetts buildings dataset, which is same-equivariant to rotation, and corresponding labels}
	\label{fig:rotation-invariance-2}
\end{figure}

Different kind of invariance to rotation transformation can be important to learn. The figure \ref{fig:rotation-invariance-1} represents the rotation invariance, where the object keeps the same label regardless its orientation. The figure \ref{fig:rotation-invariance-2} represents the same-equivariance, where the object and its label must keep the same orientation (its is common for segmentation tasks).

The main approach to learn rotation invariance is to use data augmentation to provide the CNN more examples of the same objects with different orientation. However, CNNs will often learn multiple copies of the same filter in different orientations. Some other approaches \cite{dieleman2016exploiting} try to reduce the redundancy, in order to lower the number of parameters, and thus to reduce the risk of overfitting.

\subsection{Scale invariance}

The main approach to learn scale invariance is to use data augmentation to provide the CNN more examples of the same objects with different scales. It is also possible to train several CNNs specialized in each scales and combine their predictions. Finally, other approaches consist to hard code scale invariance into the CNN such as in the Inception architecture or \cite{kanazawa2014locally}.

\chapter{Transfer Learning for Deep CNNs}

\section{Medium dataset of food (UPMC Food101)}

\subsection{Context}

\paragraph{Web API VISIIR} 
VIsual Seek for Interactive Image Retrieval (VISIIR) is a project aiming at exploring new methods for semantic image annotation. In this frame, we have contributed to the demo which uses a CNN to recognize food images across 101 categories from the Dataset UPMC\_Food101.

\paragraph{UPMC\_Food101}
It is a large multimodal dataset \cite{wang2015recipe} containing about 100,000 recipes for a total of 101 food categories. Each of them are constituted by around 800 to 950 images from Google Image found using the title of the category. Because of this, this dataset may contain some noise. It is the twin dataset of ETHZ University \cite{bossard2014food}. 

\begin{figure}
	\centering
	\includegraphics[width=.75\linewidth]{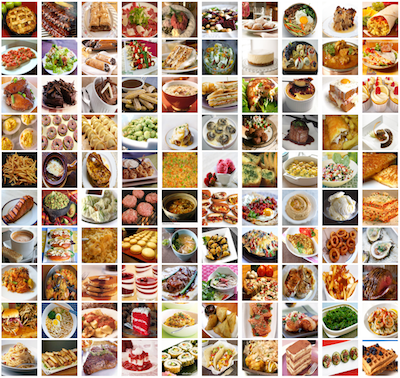}
	\caption{Illustration of the 101 food categories from the UPMC\_Food101 dataset.}
	\label{fig:2images}
\end{figure}

\subsection{Previous work}

\paragraph{CNN on UPMC\_Food101}
\cite{wang2015recipe} compared different visual features on this dataset and found that using VGG19 as features extractor was way more accurate (40.21\% top-1 accuracy) than using the BoW model with SIFT features (23.96\%). They also compared two architectures and found that a deeper architecture achieved better as features extractor than a shallower (Overfeat with 33.91\%).

\paragraph{CNN on ETHZ Food-101}
A very recent paper \cite{liu2016deepfood} claim the state of the art on two datasets of food images, UEC-256 and ETHZ Food-101. They fine tuned GoogLeNet, a 22-layer network, on these datasets and achieved a 77.4\% top-1 accuracy and 93.7\% top-5 on ETHZ Food-101.

\paragraph{Image to calories}
\cite{meyers2015im2calories} present a system aiming to recognize the contents of a meal from a single image, and then predict its nutritional contents, such as calories. They tested their CNN-based method on a dataset of 75,000 images from 23 different restaurants. Estimating the size of the foods, as well as their labels (2,516), requires solving segmentation and depth / volume estimation from a single image. They first learned a binary classifier between food and non-food class. In order to do so, they modified ETHZ Food-101 and fine tuned GoogLeNet on the new binary dataset called ETHZ Food-101 Background. They used ETHZ Food-201 Segmented to learn segmentation, and Gfood-3d to learn depth and volume. Finally they used USDA NNDB to process the amount of calorie in a certain volume of food. 
They tested each steps of there method separately and did not provide a end-to-end test. 

\subsection{Experiments}

In this subsection, our goal is to find the most accurate supervised learning method on this kind of medium sized dataset and also to understand the capacity of Convolutional Neural Networks. We use the following experimental protocol: we train each models on the same 80\% of the dataset and validate the results on the remaining 20\%. However, we select the models hyper parameters regarding our score in validation. This is the usual procedure for medium or big sized dataset (e.g. ImageNet). In table \ref{table:upmc1}, we summarize our results in order to easily compare the supervised learning methods studied. Notably, our results can be reproduced using our framework \footnote{\url{https://github.com/Cadene/torchnet-deep6/blob/master/src/main/upmcfood101/inceptionv3.lua}}.

\paragraph{Forward-backward Benchmark}
As told earlier, one of our main concern is to find the most efficient techniques to achieve the best accuracy. Since we have multi threaded the data loading and data augmentation procedure, our GPU never need to wait for the latter to finish. Thus, we make a benchmark to compare different architectures and implementations in term of speed. In this study, we use a Nvidia GTX Titan X Maxwell GPU and a Intel Xeon E5-2630 v3 (2.40GHz). We summarize our results in table \ref{table:cnnbenchmark}. Notably, InceptionV3 combined with cudnn is the fastest network. Still, it has the highest depth and number of modules. However, it has also the lowest amount of parameters. The code for this benchmark can be found online \footnote{\url{https://github.com/Cadene/torchnet-deep6/blob/master/src/main/cnnbenchmark.lua}}.

\begin{table}[h]
	\centering
	\begin{tabular}{|c|c|c|c|c|c|}
		\hline
		Model & Input size & Implementation & Forward (ms) & Backward (ms) & Total (ms) \\ \hline \hline
		Overfeat & 221 & nn float & 13714.15 & 16277.34 & 29991.49 \\ 
		Overfeat & 221 & nn cuda & 640.36 & 1032.78 & 1673.14 \\
		Overfeat & 221 & cudnn & \textbf{115.07} & 890.56 & 1005.63 \\ \hline
		Vgg16 & 224 & nn float & 18532.82 & 27978.44 & 46511.26 \\
		Vgg16 & 224 & nn cuda & 528.00 & 1407.97 & 1935.98 \\
		Vgg16 & 224 & cudnn & 458.86 & 1144.11 & 1602.96 \\ \hline
		InceptionV3 & 399 & nn float & 31934.56 & 56328.28 & 88262.84 \\
		InceptionV3 & 399 & nn cuda & 787.64 & 1312.19 & 2099.83 \\
		InceptionV3 & 399 & cudnn & 239.51 & 738.50 & \textbf{978.01} \\
		\hline
	\end{tabular}
	\caption{Benchmarks of three architectures used in our study and three implementations: nn float on CPUs (by Torch7), nn cuda on GPUs (by Torch7), cudnn R5 on Nvidia GPUs (by Nvidia). The forward and backward pass are averaged over 10 iterations. The measures are made in milliseconds for batches of size 50.} 
	\label{table:cnnbenchmark}
\end{table}

\paragraph{From Scratch better than Hand Crafted}
In table \ref{table:upmc1}, we can see that deep models trained From Scratch, (d) and (g), achieve better accuracy than Hand Crafted models, (a) and (b). In other words, it is possible to train a deep model of 140 millions parameters on a medium dataset made of 80,000 images. This might be possible, because one image is made of $224*224*3 = 150,528$ pixels. Thus, the trainset is made of 80,000 examples for 150,528 features. Also, we used data augmentation, dropout and early stopping as regularization techniques.
We virtually increase the size of the trainset as follow. Each image is randomly rescaled between 224 and 256, then randomly cropped to scale 224, and randomly flipped.
We use SGD with Nesterov momentum as our optimization technique. Especially, we use a learning rate decay to slow down the training process after each mini batch updates.
We validate the values of our hyper parameters using a unique parameters initialization (e.g. keeping the same seed).

\paragraph{From Scratch better than Features Extraction}
Deep models trained From Scratch, (d) and (g), achieve far better accuracy than models based on Features Extraction from pretrained deep models on ImageNet. Even if ImageNet contains classes and images which can be found in UPMC\_Food101, learning specialized features from scratch is more accurate. This might be possible, because the dataset is big enough and the regularization methods (e.g. data augmentation and dropout) are strong enough.

\paragraph{Fine Tuning better than From Scratch}
Fine tuned models, (e) and (h), achieve better accuracy than From Scratch models, (d) and (g). In the same manner as outlined above, it is efficient to train a deep model on this dataset, but representations learned on ImageNet, even if less accurate, can be also useful. Thus, we experimentally chose to reset all the fully connected layers of our deep networks and keep only the parameters from the convolutional layers (14,714,688 parameters in VGG16) that we update using a 10 times slower learning rate. In this manner, the networks are able to adapt their representations to UPMC\_Food101 benefiting from those learned on ImageNet.

\paragraph{Very Deep better than Deep}
Overfeat, (c), (d) and (e), has almost the same number of parameters than Vgg16, (f), (g) and (h), but achieve lower accuracy because the later is deeper (16 layers versus 9 in Overfeat). Thus, Vgg16 learn better transferable features than Overfeat. Also, the importance of depth has been experimentally shown on ImageNet.

\paragraph{InceptionV3, the most efficient architecture}
A fine tuned InceptionV3 reaches the highest accuracy on this dataset. Also, it is 3 times faster to converge than Vgg16, thanks to several factors. Firstly, it has the lowest amount of parameters. Secondly, it has no dropout layers. The latter helps to generalize, but slow down the learning process. Finally, it has batch normalization layers before each non linearity. The latter help to converge faster and also to generalize.

\begin{table}[h]
	\centering
	\begin{tabular}{|c|c|c|}
		\hline
		Model & Test top 1 (top 5) & Assoc. train top 1 (top 5)\\ \hline \hline
		(a) Bag of visual Words & 23.96 & -- \\
		(b) BossaNova & 28.59 & -- \\ \hline
		(c) Overfeat \& Extraction & 33.91 & -- \\
		(d) Overfeat \& From Scratch & 47.46 (69.37) & 79.14 (94.49) \\
		(e) Overfeat \& Fine Tuning & 57.98 (78.86) & 89.69 (97.96) \\ \hline
		(f) Vgg16 \& Extraction & 40.21 & -- \\
		(g) Vgg16 \& From Scratch & 53.62 (74.67) & 88.17 (97.68) \\
		(h) Vgg16 \& Fine Tuning & 65.71 (82.54) & 96.18 (99.39) \\ \hline
		(g) InceptionV3 \& Fine Tuning & \textbf{66.83 (84.53)} & 85.34 (95.91) \\
		\hline
	\end{tabular}
	\caption{Percentage of good classification on UPMC\_Food101. (a), (b), (c) and (f) are reported from \cite{wang2015recipe}}
	\label{table:upmc1}
\end{table}

\section{Small dataset of objects (VOC2007)}

\subsection{Context}

\paragraph{PASCAL VOC 2007 dataset} The goal of the PASCAL Visual Object Classes Challenge 2007 is to recognize objects from 20 visual object classes in realistic scenes (i.e. not pre-segmented objects). It is fundamentally a multi-label supervised learning problem in that a training set of labelled images is provided (5,000 images compose the training set and 5,000 images compose the testing set). The twenty object classes that have been selected are:
\begin{itemize}
	\item Person: person,
	\item Animal: bird, cat, cow, dog, horse, sheep,
	\item Vehicle: aeroplane, bicycle, boat, bus, car, motorbike, train,
	\item Indoor: bottle, chair, dining table, potted plant, sofa, tv/monitor
\end{itemize}

\begin{figure}[h]
	\centering
	\includegraphics[width=.99\linewidth]{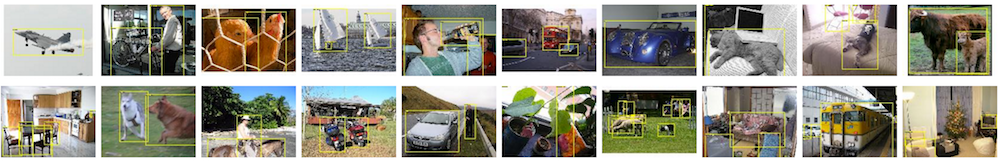}
	\caption{Illustration of the 20 object classes from the PASCAL\_VOC2007 dataset.}
	\label{fig:2images}
\end{figure}

\subsection{Previous work}

\begin{figure}[h]
	\begin{subfigure}{.50\textwidth}
		\centering
		\includegraphics[width=1\linewidth]{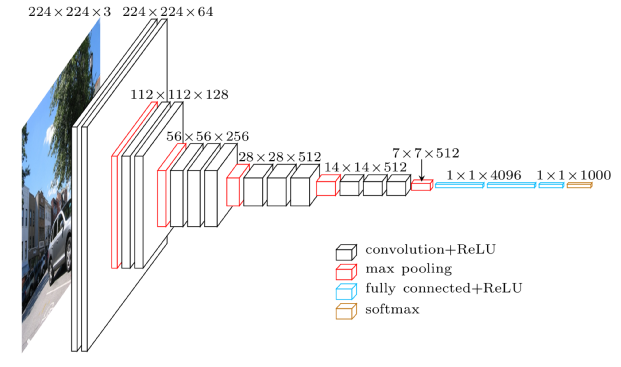}
		\caption{Convolutional Network}
		\label{fig:sfig1}
	\end{subfigure}%
	\begin{subfigure}{.50\textwidth}
		\centering
		\includegraphics[width=.95\linewidth]{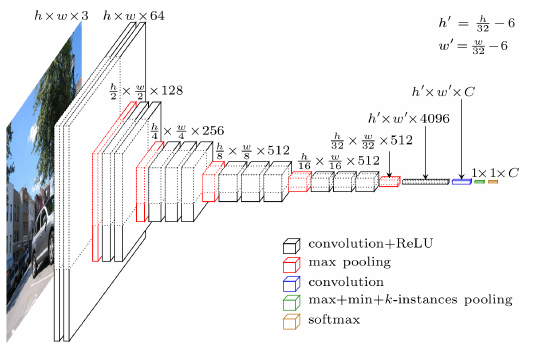}
		\caption{Fully Convolutional Network}
		\label{fig:sfig2}
	\end{subfigure}
	\caption{Illustration of a possible modification of VGG16 to a fully convolutional architecture in order to process bigger image than of size 3x224x224.}
	\label{fig:CNNfullyconvolutional}
\end{figure}

\paragraph{Dense Testing}
\cite{simonyan2014very} evaluated the generalization capacity of their CNNs, namely VGG16 and VGG19, on VOC-2007, VOC-2012, Caltech-101 and Caltech-256. They proposed a method called dense testing. The network is applied densely over the rescaled test images, i.e. the fully-connected layers are first converted to convolutional layers (the first FC layer to a 7 x 7 convolutional layer, the last two FC layers to 1 x 1 convolutional layer). The resulting fully-convolutional net is then applied to the whole (uncropped) image (see figure \ref{fig:CNNfullyconvolutional}). The result is a class score map with the number of channels equal to the number of classes. Then, to obtain a fixed-size vector of class scores for the image, the class score map is spatially averaged (Average Pooling). The test set is also augmented by horizontal flipping of the images. Finally, the soft-max class posteriors of the original and flipped images are averaged to obtain the final scores for the image.

They combined three CNNs fine tuned at different scales on ImageNet and achieved an impressive score of 89.7 mean AP on VOC-2007 and 89.3 on VOC-2012. Using the same amount of data, the highest score reported at this time was 82.4 on VOC-2007 and 83.2 on VOC-2012.

\subsection{Experiments}


\paragraph{Fine tuning outperforms the rest} As can bee seen in table \ref{table:voc1}, deep features outperform hand-crafted methods and fine tuning Vgg16 pretrained on ImageNet achieves the best accuracy. However the dataset is too small to fit Vgg16 From Scratch. In this experiment, we train our networks on multiple scales varying from 224 to 256 with random horizontal flips, so that the forwarded cropped image is of size $224 \times 224$. Then, we test our models on single scale images (224). In (e), we reset the last two fully connected layers and we train the others with a 10 times smaller learning rate. In (d), we extract deep features from the last ReLU non linearity (before the last fully connected layer) and then we train support vector machines in a one-versus-all strategy cross-validating the regularization parameter.

\begin{table}[H]
	\centering
	\begin{tabular}{|c|c|c|}
		\hline 
		Model type & Test mAP & Train mAP\\ \hline \hline
		(a) BoW & 53.2 & -- \\
		(b) BossaNova and FishersVector & 61.6 & -- \\
		\hline
		(c) Vgg16 from scratch & 39.79 & 99.73 \\
		(d) Vgg16 extraction & 83.22 & -- \\
		(e) Vgg16 fine tuned & \textbf{85.70} & 98.81 \\
		\hline
	\end{tabular}
	\caption{Comparison between hand crafted features models (a,b) and deep features models (c,d,e) tested on a single scale (224). (a) and (b) are reported from \cite{avila2013pooling}.}
	\label{table:voc1}
\end{table}

\section{Small dataset of roofs (DSG2016 online)}

In this section, we show that fine tuning an InceptionV3 architecture is still the most powerful technique for this kind of dataset. However, we use a bootstrap method to reduce overfitting and thus to achieve the best accuracy on this dataset amongst more than 110 teams.

\subsection{Context}

\paragraph{Data Science Game}
The DSG \footnote{\url{http://www.datasciencegame.com}} is an international student challenge in Machine Learning. The first phase of this challenge is an online selection. It took place on Kaggle from the 14th June 2016 until the 10th July 2016 gathering more than 110 teams from all around the world competing for the first 20 positions. The final phase will took place during the week end of the 10th September in the castle of Cap Gemini. 

The team (Jonquille UPMC) that we represented reached the first position of the online selection \footnote{\url{https://inclass.kaggle.com/c/data-science-game-2016-online-selection/leaderboard}}. In this section, we will explain our method in details.

\paragraph{Dataset}
It is a small dataset of satellite images of roofs. The goal is to predict the orientation of the roofs into 4 different categories. Models are evaluated with the multiclass accuracy top1 metric. The public leaderboard is made of 40\% of the testing set. The private leaderboard is made of 60\% others.
The dataset is composed by:
\begin{itemize}
	\item 8000 images in the training set, 3479 of which are in the category \textit{North-South orientation}, 1856 in the category \textit{East-West orientation}, 859 in the category \textit{Flat roof} and 1806 in the category \textit{Other},
	\item 20,760 images in the training set, but without any label,
	\item 13,999 images in the testing set.
\end{itemize}

\begin{figure}
	\centering
	\begin{subfigure}{.23\textwidth}
		\centering
		\includegraphics[width=.75\linewidth]{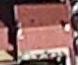}
		\caption{North-South}
		\label{fig:dsg1}
	\end{subfigure}%
	\begin{subfigure}{.23\textwidth}
		\centering
		\includegraphics[width=.75\linewidth]{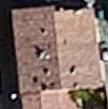}
		\caption{East-West}
		\label{fig:dsg2}
	\end{subfigure}
	\begin{subfigure}{.23\textwidth}
		\centering
		\includegraphics[width=.75\linewidth]{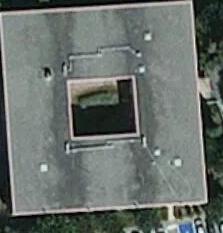}
		\caption{Flat}
		\label{fig:dsg3}
	\end{subfigure}
	\begin{subfigure}{.23\textwidth}
		\centering
		\includegraphics[width=.75\linewidth]{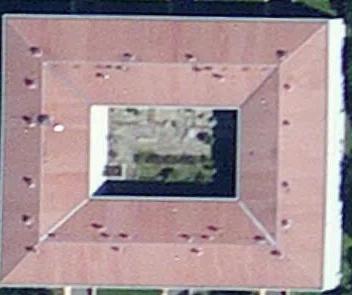}
		\caption{Other}
		\label{fig:dsg4}
	\end{subfigure}
	\caption{Illustrations of the DSG2016 online selection dataset}
	\label{fig:dsgimages}
\end{figure}

\paragraph{Rules} Pretrain models are usable only if they are publicly available and if they are not trained on roof datasets.

\subsection{Experiments}

\paragraph{Winning solution}
One of the main difficulties of this challenge was the relatively small size of the dataset. Yet, we thought that the latter was big enough to fine-tune a pretrained model on. Our very first submission was a VGG16 finetuned on 80\% of the trainset with some data augmentation and validated on the rest. It scored a whooping 82.67\% on the public leaderboard (and 81.37\% on the private), which put us on the top 3 at the time, and confirmed it was working.

We then tried InceptionV3 because of its good performance in many transfer learning setups. In order to reduce overfitting, we used the following techniques: online data augmentation, the 90° trick (multiplying by 2 the number of images and moving the rotated images from class "north-south" to class "east-west" and vice-versa) and early stopping. We didn't add more features (images height / weight) because it seemed to lead to more overfitting, which was a main concern.

Still, it was difficult to evaluate our models on the little test data that remained. There was a lot of variance between different epochs, folds and hyperparameters. To gain more stability and to use the full training set, we went to the bagging method. We fine tuned some InceptionV3 with different hyperparameters on random bootstraps. The final prediction is the result of a vote among all classifiers.

We also tweaked the predictions of our final submission to counteract the fact that the train set had unbalanced classes, while the test set was balanced. This improved a bit our score.

\paragraph{Other solutions}
We fine tuned in a end-to-end manner a Spatial Transformer (STN) architecture with a first InceptionV3 to localize a region of interest and a second InceptionV3 to classify this region. We tried to constraint the kind of spatial transformation the network could achieve, or the number of parameters to tune. However, we think that the localizer learned the identity transformation. Also, it was expensive to train this network (twice the number of parameters compared to a simple InceptionV3) and overfitting was a main concern, thus we stopped to explore this solution.

\paragraph{Second team solution}
The team ranked second used a semi-supervised technique to augment their training set with unlabeled images. They also used a stacking of 84 models: 12 root models trained on 7 different versions of the images \footnote{\url{https://medium.com/@Zelros/how-deep-learning-solved-phase-1-of-the-data-science-game-2712b949963f}}. However they seemed to have overfitted too much the public testing set regarding the gap with their private score.

\begin{table}[H]
	\small
	\centering
	\begin{tabular}{|c|c|c|c|c|}
		\hline
		Fine tuned models & \# Models & Data & Public (\%) & Private (\%) \\ \hline \hline
		VGG16 & 1 &80\%trainset & 82.67 & 81.37 \\
		InceptionV3 & 1 &80\%trainset & 83.69 & 82.61\\
		InceptionV3 + STN & 1 &80\%trainset & 84.73 & 84.11\\		
		VGG19 (Second Team) &84& 100\%trainset + no label & \textbf{87.60} & 86.38 \\
		InceptionV3 & 91 &100\%trainset & 87.32 & 86.58\\
		InceptionV3 + Prior & 91 & 100\%trainset & 87.52 & \textbf{86.76}\\
		\hline
	\end{tabular}
	\caption{Summary of the different models submitted for the DSG online challenge.}
	\label{table:voc3}
\end{table}

\section{Conclusion}

In this chapter, we studied the transfer capacity of the latest convolutional architectures. To do so, we compared three training approaches on several datasets which all have their own particularity : size, semantic distance from ImageNet, geometric variability of their regions of interest.

Recall that our approaches can be synthesize as follow: 
\begin{itemize}
	\item We trained specific deep convolutional networks randomly initializing their parameters (e.g. From Scratch).
	\item Further, we used pre-trained networks on ImageNet to extract features from the target dataset and trained a linear model (e.g. Features Extraction).
	\item Finally, we fine tuned previous pre-trained networks to adapt their learned representations to the target dataset (e.g. Fine Tuning).
\end{itemize}

Each datasets represent a different challenge. In term of accuracy, we illustrated few trends:
\begin{itemize}
	\item Fine Tuning achieve the best accuracy on medium datasets, and From Scratch achieves better accuracy than Features Extraction.
	\item Fine Tuning still achieve the best accuracy on small datasets, but From Scratch achieves lower accuracy than Features Extraction. Also, approaches based on a bagging of models are effective to achieve better accuracy.
\end{itemize}

In term of learning ability, we illustrated few other trends:
\begin{itemize}
	\item InceptionV3 is the most accurate architecture and its batch normalization layers allow to converge faster.
	\item Adam is useful to reduce the number of experiments needed to find the optimal set of hyper parameters.	
	\item Early stopping is a good way to control overfitting.
\end{itemize}

\chapter{Weakly Supervised Learning}

\section{Introduction}

\subsection{Definition}

We define Weakly Supervised Learning (WSL) as a machine learning framework where the model is trained using examples that are only partially annotated or labeled. For instance, an object detector is typically trained on large collection of images manually annotated with masks or bounding boxes denoting the locations of objects of interest in each image. The reliance on time-consuming human labeling poses a significant limitation to the practical application of these methods. Moreover, manually annotation may not be optimal for the final prediction task. Thus, WSL aims at reducing the amount of human intervention needed.

Despite their excellent performances, current CNN architectures only carry limited invariance properties. Recently, attempts have been made to overcome this limitation using WSL. Many different kind of WSL techniques exists in the literature according to the type of labeled data used, latent variables learned and evaluation methods selected. We also consider that certain papers on multiple instance learning, attention based models or fine grained classification belong to a larger WSL framework. Below, we illustrate our thinking.

\subsection{Multi Instance Learning}


\paragraph{Weakly Supervised Learning (WSL) with Max Pooling}
In computer vision, the dominant approach for WSL is the Multiple Instance Learning (MIL) paradigm. An image is considered as a bag of regions, and the model seeks the max scoring instance in each bag. In \cite{oquab2015object}, the authors fixed the weights of a Vgg16 network and learned the last fully connected layer (e.g. 1x1 convolutional layer). They applied Vgg16 to images of bigger size than the original input (224x224). The resulting class score map is spatially reduced using a Max Pooling. They proposed two methods to learn scale invariance. The first was to rescale randomly the input image. The second was to train three specialized CNNs at three different scales and then to average their class scores. The two methods lead to almost the same results. They achieved 86.3 mean AP on VOC-2012. To compare, the same CNN used as a classical features extractor achieved 78.7 mean AP.

\begin{figure}[h]
	\centering
	\includegraphics[width=.40\linewidth]{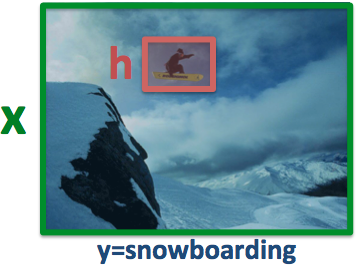}
	\caption{Illustration of a Weakly Supervised Learning approach which learn to focus on a region of interest from a global label.}
	\label{fig:2images}
\end{figure}

\begin{figure}[h]
	\centering
	\includegraphics[width=.60\linewidth]{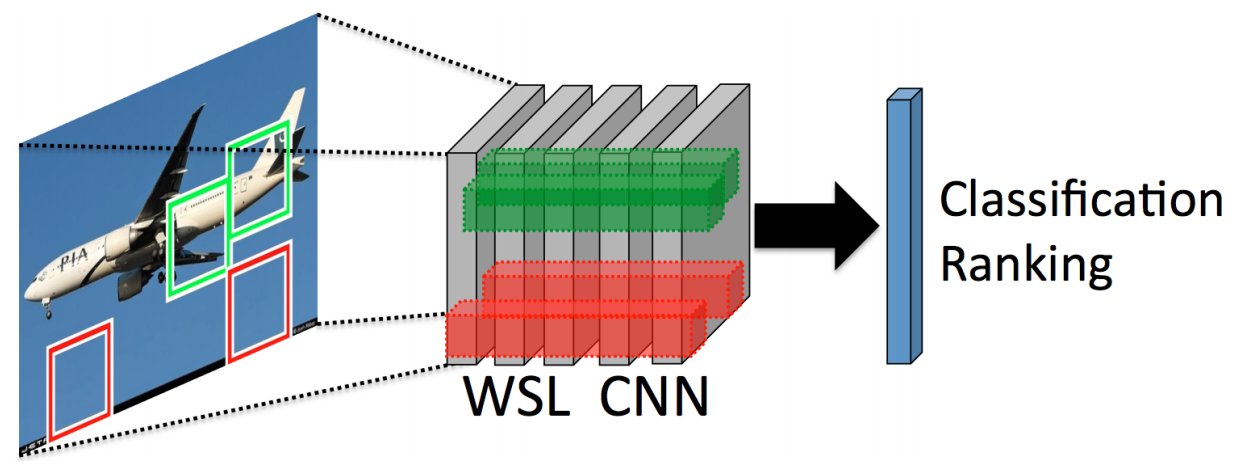}
	\caption{Illustration of the WELDON approach, a CNN trained in a weakly supervised manner to perform classification or ranking. It automatically selects multiple positive (green) or negative (red) evidences on several regions in the image.}
	\label{fig:Weldon_figure1}
\end{figure}

\paragraph{WSL with k Max Min Pooling (WELDON)}
In \cite{durand2016weldon}, the authors proposed to use the same method than in \cite{oquab2015object}, but the class score map is spatially reduced using a sum of k-Min Pooling and k-Max Pooling. As a result, it extends the selection of a single region to multiple high score regions incorporating also the low score regions, i.e. the negative evidences of a class appearance.
They also proposed a ranking loss to optimize average precision.
Using an average of 7 specialized CNNs at 7 different scales, they achieved 90.2 mean AP on VOC-2007 and 88.5 on VOC-2012.
They also claim the state of the art on 5 other datasets. See figure \ref{fig:Weldon_figure1}.

\subsection{Spatial Transformer Network}

\begin{figure}
	\centering
	\includegraphics[width=.85\linewidth]{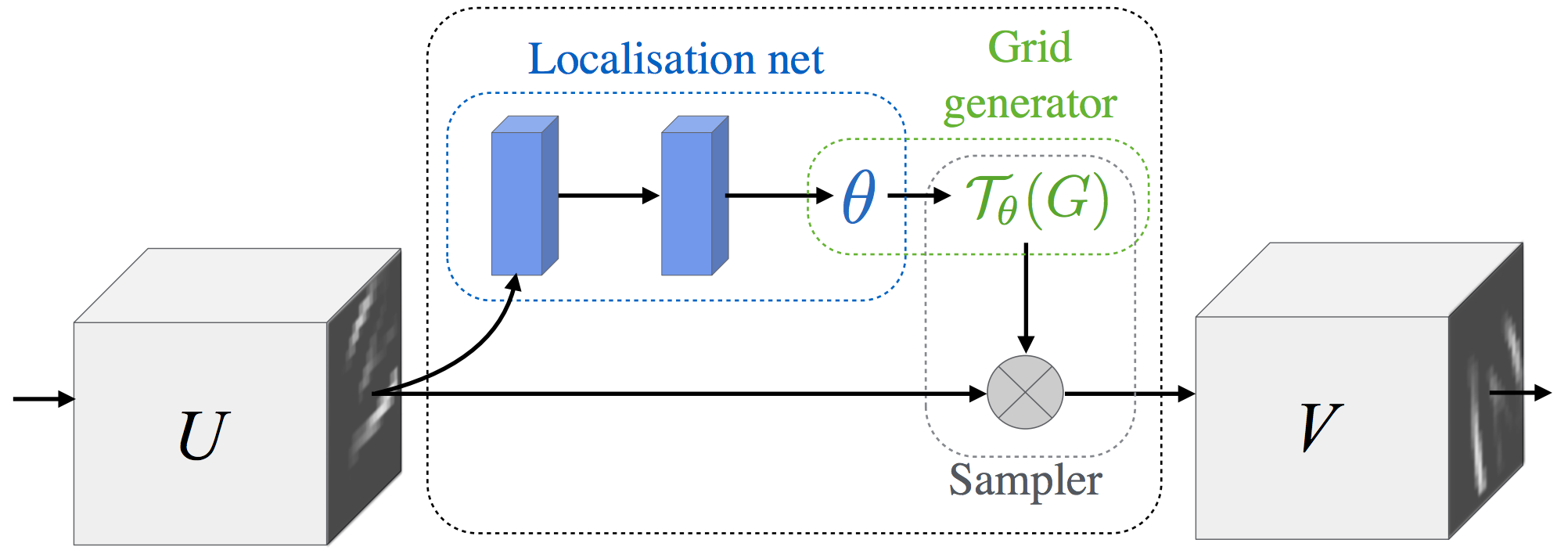}
	\caption{Spatial Transformer Module. The input feature map U is passed to a localization network which regresses the transformation parameters $\theta$. The regular spatial grid $G$ over $V$ is transformed to the sampling grid $\mathcal{T}_\theta (G)$, which is applied to $U$, producting the warped output feature map $V$.}
	\label{fig:STN}
\end{figure}

\paragraph{Principles}
The authors of \cite{jaderberg2015spatial} introduce a new learnable module, the Spatial Transformer, which explicitly allows the spatial manipulation of data within the network. This differentiable module can be viewed as a localization network which generate parameters for a grid generator. The grid is then used by a sampler to generate a certain transformation. The latter can be applied on the initial image or on the same feature map used as input to the localization network such as in figure \ref{fig:STN}.

\paragraph{Transformations}
Multiple transformation can be learned. For instance, $\mathcal{T}_\theta (G)$ can apply a 2D affine transformation $A_\theta$. In this case, the pointwise transformation is
$$
\left( \begin{array}{ c } x^s_i \\ y^s_i \end{array} \right) =
 \mathcal{T}_\theta (G) =
 A_\theta \left( \begin{array}{ c } x^t_i \\ y^t_i \\ 1 \end{array} \right) =
 \left[\begin{array}{ c  c  c }
 \theta_{11} & \theta_{12} & \theta_{13} \\
 \theta_{21} & \theta_{22} & \theta_{23} 
 \end{array} \right] 
\left( \begin{array}{ c } x^t_i \\ y^t_i \\ 1 \end{array} \right) 
$$

However the class of transformations $\mathcal{T}_\theta$ may be more constrained, such as that used for attention
$$
 A_\theta =  \left[\begin{array}{ c  c  c }
s & 0 & t_x \\
0 & s & t_y
 \end{array} \right] 
 $$
allowing cropping, translation, and isotropic scaling by varying $s$, $t_x$, and $t_y$. Those parameters can be learned or, to constrain even more, fixed.

\paragraph{Traffic signs classification}

The first example of application is the supervised classification of a traffic signs dataset (GTSRB \footnote{\url{http://benchmark.ini.rub.de/?section=gtsrb&subsection=news}}). This kind of data is affected by contrast variation, rotational and translational changes. \cite{haloi2015traffic} used a spatial transformer network to make classification more robust and accurate. They used a modified version of GoogLeNet with batch normalization and added four different Spatial Transformer module made of a convolutional network as localizer. 
This method has several advantages over existing state of the art methods in terms of performance, scalability and memory requirement. Also,  they reported a lower accuracy for their architecture without spatial transformer modules (99.57\% against 99.81\%). An implementation is available on the Torch7 blog \footnote{\url{http://torch.ch/blog/2015/09/07/spatial_transformers.html}}.

\begin{figure}[h]
	\centering
	\includegraphics[width=.50\linewidth]{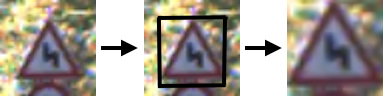}
	\caption{The use of a Spatial Transformer Module on a GTSRB data sample.}
	\label{fig:STN_GTSRB}
\end{figure}

\paragraph{Fine grained bird classification}

The second example is the fine grained classification of a bird dataset (CUB-200-2011). The birds appear at a range of scales and orientations, are not tightly cropped, and require detailed texture and shape analysis to distinguish. This dataset contains 6k training images, 5.8k test images and covers 200 species of birds. They first claimed the state of the art accuracy of 82.3\% with an Inception architecture pre-trained on ImageNet and fine tuned on CUB (previous best was 81.0\%). Then, they trained a spatial transformer network containing 4 parallel spatial transformer modules parameterised for attention and acting on the input image. They achieved an accuracy of 84.1\% outperforming their baseline by 1.8\%. The resuling output from the spatial transformers for the classification network is somewhat pose-normalised representation of a bird. It was able to discover and learn part detectors in a data-driven manner without any additional supervision. Also, the use of spatial transformers allowed them to use 448px resolution input images without any import in performance as the output of the transformed 448px images are downsampled to 224px before being processed.

\begin{figure}[h]
	\centering
	\includegraphics[width=.95\linewidth]{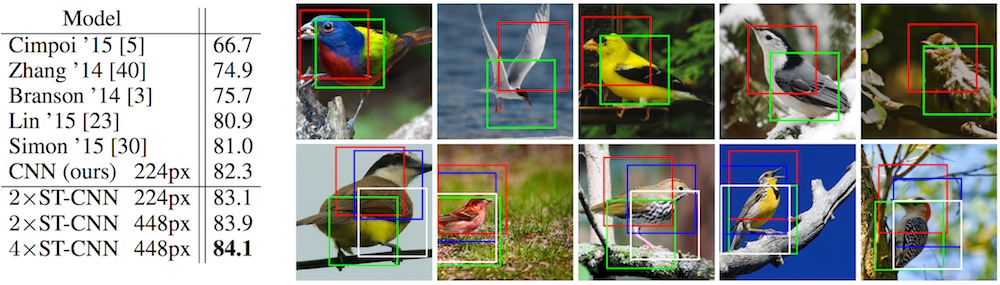}
	\caption{Left: the accuracy on CUB-200-2011 bird classification dataset. Spatial transformer networks with two spatial transformers modules (2 x ST-CNN) and four (4 x ST-CNN) in parallel achieve higher accuracy. Right: the transformation predicted by the 2 x ST-CNN (top row) and 4 x ST-CNN (bottom row) on the input image. Notably, for the 2 x ST-CNN, one of the module (shown in red) learned to detect heads, while the other (shown in green) detects the body.}
	\label{fig:STN_GTSRB}
\end{figure}

\section{Applying fine tuning to Weldon}

\subsection{Context}

\paragraph{MIT67} It is the database of the Indoor scene recognition challenge \footnote{\url{http://web.mit.edu/torralba/www/indoor.html}}. It was released during the CVPR 2009 \cite{quattoni2009recognizing}. It contains 67 categories. 5360 images compose the training set. 1340 images compose the testing set. The number of images does not vary across categories. This dataset is quite similar to Pascal Voc in term of size and difficulty, e.g. while some indoor scenes (e.g. corridors) can be well characterized by global spatial properties, others (e.g., bookstores) are better characterized by the objects they contain.

\begin{figure}[h]
	\centering
	\includegraphics[width=1.\linewidth]{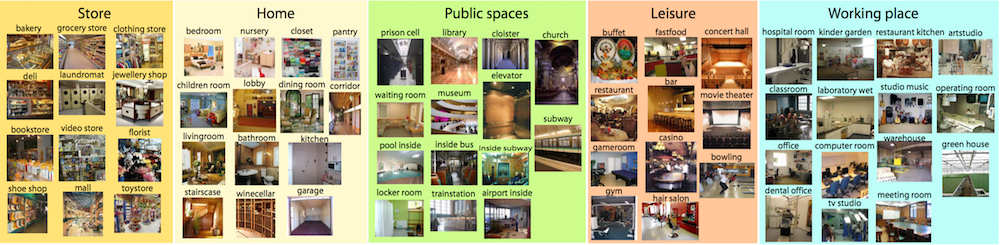}
	\caption{Illustration of the 67 indoor categories from the MIT67 dataset.}
	\label{fig:mit67}
\end{figure}



\subsection{Experiments}

\paragraph{Fine tuning a Weldon architecture}  We reproduce the results of \cite{durand2016weldon} and show in table \ref{table:weldon1} that Fine Tuning the Weldon architecture at three different scales improves the overall accuracy. However, the process of Fine Tuning is expensive, thus we do not provide results for bigger scale.

\begin{table}[H]
	\small
	\centering
	\begin{tabular}{|c|c|c|c||c|c|}
		\hline
		Model &Image size & L6 size & Report (*) & Extraction & Fine Tuning \\ \hline \hline
		Vgg16 & $224 \times 224$ & $1 \times 1$ & 69.93 & 70.30 & 70.60 \\
		\hline
		Weldon & $249 \times 249$ & $2 \times 2$ & 72.16 & 72.01 & 73.4 \\
		Weldon & $280 \times 280$ & $3 \times 3$ & 72.98 & 73.80 & 74.03 \\
		Weldon & $320 \times 320$ & $4 \times 4$ & 73.40 & 73.96 & 74.1\\
		\hline
		\end{tabular}
		\caption{Accuracy top 1 on MIT67 test set. Without data augmentation and dropout. $k=1$ aggregation (one region min, one region max). L6 size represents the number of regions (e.g. instances) evaluated during the aggregation. For size 320, 16 regions are evaluated. Results from column Report (*) are reported from \cite{durand2016weldon}. }
		\label{table:weldon1}
\end{table}

\section{Study of Spatial Transformer Network}

\subsection{Context}

\paragraph{Database}
In this section, we study the ability of the Spatial Transformer Network to learn large spatial invariance. In order to do so we create a special dataset using the original MNIST dataset padded with 2 black pixels (e.g. all our images are of scale $32\times 32$ instead of the original $28\times 28$). Then, we create a Translated MNIST dataset. All the images are inserted on a $100\times 100$ background full of black pixels, and undergo spatial shifts on the x and y axes. Those shifts are randomly picked up between 0 and 68 (=100-32). 

\subsection{Previous work}

\paragraph{Co-localization}
In the appendix A.2 of \cite{jaderberg2015spatial}, the authors explored the use of spatial transformers in a co-localization scenario. Given a set of images that are assumed to contain instances of a common but unknown object class, the model learn from the images only to localize (with a bounding box) the common object. To achieve this, they adopted the supervision that the distance between the image crop corresponding to two correctly localized objects is smaller than to a randomly sampled image crop, in some embedding space. For a dataset  $\mathcal{I} = {I_n}$ of $N$ images, this translates to a triplet loss, where they minimized the hinge loss 
$$
\sum^N_n \sum^M_{m \neq n} max (0, ||e(I^\mathcal{T}_n) - e(I^\mathcal{T}_m) ||^2_2 - ||e(I^\mathcal{T}_n) - e(I^{rand}_n) ||^2_2 + \alpha)
$$

They used translated (T), and translated and cluttered (TC) MNIST images (28x28) on a 84x84 black background. As features extractor ($e(.)$), they used a pretrain network on MNIST. As localizer, they used a 100k parameters CNN. Also, they used a spatial transformer module parameterized for attention (scale, translation, no rotation). 

They measured a digit to be correctly localized if the overlap (are of intersection divided by area of union) between the predicted bounding box and groundthruth bounding box is greater than 0.5. On T they got 100\% accuracy. On CT between  75-93\%.

\begin{figure}[h]
	\centering
	\includegraphics[width=1.\linewidth]{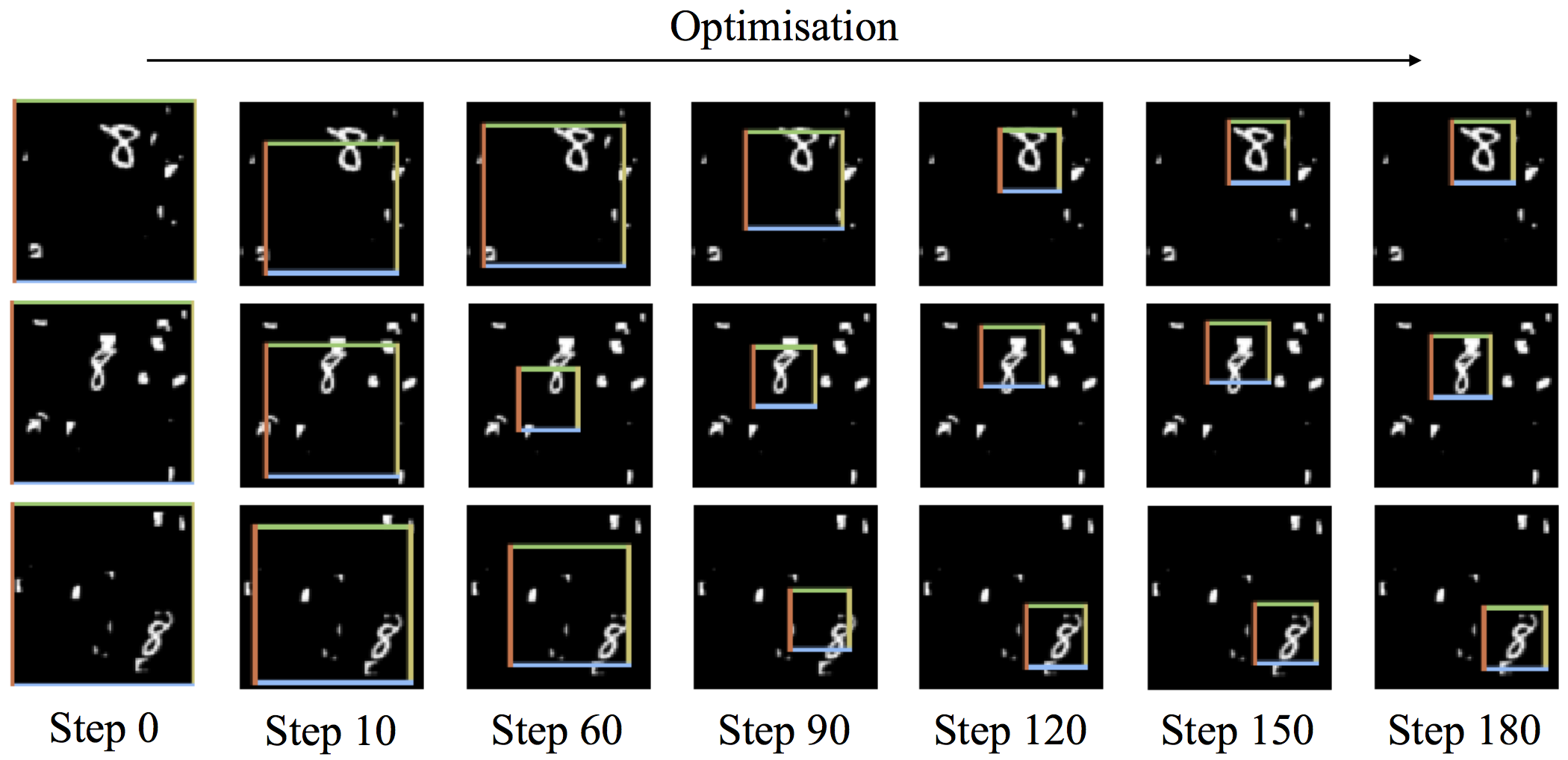}
	\caption{Illustration of the dynamics for co-localization. Here are the localization predicted
		by the spatial transformer for three of the 100 dataset images after the SGD step labelled below. By SGD
		step 180 the model has process has correctly localized the three digits.}
	\label{fig:mit67}
\end{figure}


\subsection{Experiments}

Almost all our experiments are made with a batch size of 256, Adam as optimizer, a learning rate of $3e^{-4}$,  a learning rate decay of 0 and a weight decay of 0.


On table \ref{table:STN_bench}, we compare several approaches to a toy problem. The goal for a STN is to localize where the white pixels on the initial images are, then to generate a zoomed-in view of the digit.

\paragraph{ConvNet abilities to learn strong spatial invariance.} 
(a) uses a LeNet-5 as classifier. It is an optimized architecture for the classification of 28x28 digits on a $32\times 32$ background.
In this case, LeNet-5 takes as input downsampled images. This process reduces the readability of digits. Thus, LeNet-5 is only able to achieve a 85.07\% accuracy.

(b) uses a bigger convolutional network (ConvNet100).
We want to measure the cost of the downsampling process. Thus, this network takes as input images of size $100\times 100$. Finally, it is able to achieve a 96.12\%, 11\%  more than LeNet-5 (a). However, we do not use any data augmentation procedure.

\paragraph{STN abilities}
(c,d,e,g) uses a LeNet-5 in the same way and apply a transform to the initial image ($100\times 100$). However, the latter is also downsampled from $100\times 100$ to $32\times 32$ to fit the input size of the localizer.

Firstly, we show that even when using bad resolution images, the localizer is able to produce good spatial transformations, leading to almost the same accuracy as using the full resolution (d, f, i). Secondly, we show that generating 3, 4 or 6 parameters lead to almost the same accuracy.

\paragraph{Multi Instance Learning (MIL) abilities}
(f) uses a WSL with Max Pooling architecture as described in subsection 3.1.2. LeNet-5 is transformed to a fully convolutional network in order to take as input images of size $100\times 100$. The final features map is spatially aggregated by a Max Pooling of the size of the features map (e.g. the output is of size 1x1x10). This method is the fastest to converge and lead to one of the highest accuracy (99.15\%). We have the same results fine tuning a pretrain LeNet-5 on the original MNIST (e.g. a background of size $32\times 32$) with an accuracy of 99.05\%.

\begin{table}[H]
	\small
	\centering
	\begin{tabular}{|c|c|c|}
		\hline
		Model & Input size & Test accuracy top1 \\ \hline \hline

		(a) LeNet-5 & $32\times32$ & 85.07 \\
		(b) ConvNet100 & $100\times100$ & 96.12\\
		(c) STN affine & $32\times32$ & 99.06 \\
		(d) STN translation & $32\times32$ & 99.10\\
		(e) STN translation+scale & $32\times32$ &  99.10 \\	
		(f) MIL MaxPooling & $100\times100$ & 99.15 \\			
		(g) STN translation+scale+rotation & $32\times32$ & 99.18 \\
		\hline
	\end{tabular}
	\caption{Comparison of different methods applied on our MNIST Translated dataset with a background of size $100\times 100$. We estimate that our results may vary plus or minus 0.10.  }
	\label{table:STN_bench}
\end{table}

\section{Conclusion}

In this chapter, we studied Weakly Supervised Learning (WSL) approaches which can be synthesized as follow:
\begin{itemize}
	\item Multi Instance Learning (MIL) with Max Pooling which considers an image as a bag of regions and seeks the max scoring region.
	\item MIL with k Max Min Pooling (e.g. Weldon) which extends the selection of a single region to multiple high score regions and low score regions.
	\item Spatial Transformer Network (STN) which uses a network (e.g. localizer) that takes as input the original image and generates a transformed image. Thus, the second network (e.g. classifier) takes as input a invariant representation of the object to classify.
\end{itemize}

In a first section, we studied the Weldon approach on a small and complex dataset (e.g. MIT67).
\begin{itemize}
	\item We explained that it is well suited for this kind of datasets where regions of interest are multiple and negative evidences of classes are present.
	\item Especially, we showed that fine tuning is well suited for Weldon architectures.
\end{itemize} 

In a second section, we studied the STN approach.
\begin{itemize}
	\item Firstly, we explained in which cases STN are used and in which forms.
	\item Secondly, we compared on a toy dataset (e.g. Translated MNIST) classical Convolutional Neural Networks (CNNs), STN and MIL with Max Pooling approaches in term of spatial invariance capacity. Thus, we showed that STN was able to generate invariant representations of the digits in order to achieve better accuracy than CNNs and the same accuracy than MIL.
\end{itemize}

\chapter*{Conclusion}
\addcontentsline{toc}{chapter}{Conclusion}

\paragraph{Summary of Contributions}
In this master's thesis, we studied deep learning architectures for classifying medium and small datasets of images.
\begin{itemize}
\item In a first chapter, we explained how Convolutional Neural Networks can achieve such good accuracy.
\item In a second chapter, we showed the efficiency of the Fine Tuning approach on this kind of dataset. We also explained our winning solution to the DSG online challenge based on a bootstrap of fine tuned InceptionV3. 
\item In a last chapter, we showed the advantages and drawbacks of Weakly Supervised Learning approaches such as Multi Instance Learning (MIL) and Spatial Transformer Networks (STN). Using Fine Tuning, we also improved Weldon, a certain kind of MIL model.
\end{itemize}

\paragraph{Future Directions}
 In future studies, we would like to adapt the methods developed during this study to multi-modal datasets made of images and texts. Furthermore, we would like to apply Fine Tuning and Weakly Supervised Learning on the last architectures such as Wide Residual Networks. Finally, we would like to explore hybrid weakly supervised architectures counteracting the drawbacks of MIL and STN, and seeking improvements.

\begin{appendices}

\chapter{Overfeat}

\begin{table}[h]
	\centering
	\resizebox{380pt}{!}{%
		\begin{tabular}{|c|c|c|}
			\hline
			Layer id & Layer type & Parameters number \\ \hline \hline
			(0): & Image (3, 221, 221) & \\
			(1): & Convolution (96, 3x7x7, 2x2, 0x0) & 14,208 \\
			(3): & MaxPooling (3x3,3x3) & \\
			(4): & Convolution (256, 96x3x3, 7x7) & 1,204,480 \\
			(6): & MaxPooling (2x2,2x2) & \\
			(7): & Convolution (512, 256x3x3, 1x1, 1x1) & 1,180,160\\
			(9): & Convolution (512, 512x3x3, 1x1, 1x1) & 2,359,808	 \\
			(11): & Convolution (1024, 512x3x3, 1x1, 1x1) & 4,719,616	 \\
			(13): & Convolution (1024, 1024x3x3, 1x1, 1x1) & 9,438,208	 \\
			(15): & MaxPooling (3x3,3x3) & \\
			(16): & FullyConnected (25600 $\rightarrow$ 4096) & 104,861,696 \\
			(18): & FullyConnected (4096 $\rightarrow$ 4096) & 16,781,312 \\
			(20): & FullyConnected (4096 $\rightarrow$ 1000) & 4,097,000 \\
			(21): & SoftMax & \\ \hline
			Total : & \textbf{9 layers} & \textbf{144,656,488} \\
			\hline
		\end{tabular}}
		\caption{Deep architecture used in our experiments. A ReLU non-linearity follows each convolutional and fully-connected layers, beside the last one.
			Convolution (512, 512x3x3, 1x1, 1x1) means 512 filters (e.g. 512 output channels), a kernel size of 256x3x3, 1 step of the convolution to the width and height dimensions, 1 additional zero padded per width to the input, 1 per hight.
			MaxPooling (2,2,2,2) means a 2D pooling operation on 2x2 pixels neighborhood, by step size of 2x2}
		\label{table:Overfeat}
	\end{table}
	
\chapter{Vgg16}

\begin{table}[h]
	\centering
	\resizebox{242pt}{!}{%
	\begin{tabular}{|c|c|c|}
		\hline
		Layer id & Layer type & Parameters number \\ \hline \hline
		(0): & Image (3, 224, 224) & \\
		(1): & Convolution (64, 3x3x3, 1x1, 1x1) & 1,792 \\
		(3): & Convolution (64, 64x3x3, 1x1, 1x1) & 36,928 \\
		(5): & MaxPooling (2x2,2x2) & \\
		(6): & Convolution (128, 64x3x3, 1x1, 1x1) & 73,856 \\
		(8): & Convolution (128, 128x3x3, 1x1, 1x1) & 147,584 \\
		(10): & MaxPooling (2x2,2x2) & \\
		(11): & Convolution (256, 128x3x3, 1x1, 1x1) & 295,168 \\
		(13): & Convolution (256, 256x3x3, 1x1, 1x1) & 590,080 \\
		(15): & Convolution (256, 256x3x3, 1x1, 1x1) & 590,080 \\
		(17): & MaxPooling (2x2,2x2) & \\
		(18): & Convolution (512, 256x3x3, 1x1, 1x1) & 1,180,160 \\
		(20): & Convolution (512, 512x3x3, 1x1, 1x1) & 2,359,808 \\
		(22): & Convolution (512, 512x3x3, 1x1, 1x1) & 2,359,808 \\
		(24): & MaxPooling (2x2,2x2) & \\
		(25): & Convolution (512, 512x3x3, 1x1, 1x1) & 2,359,808 \\
		(27): & Convolution (512, 512x3x3, 1x1, 1x1) & 2,359,808 \\
		(29): & Convolution (512, 512x3x3, 1x1, 1x1) & 2,359,808 \\
		(31): & MaxPooling (2x2,2x2) & \\
		(32): & FullyConnected (25088 $\rightarrow$ 4096) & 102,764,544 \\
		(34): & Dropout (50\%) & \\
		(35): & FullyConnected (4096 $\rightarrow$ 4096) & 16,781,312 \\
		(37): & Dropout (50\%) & \\
		(38): & FullyConnected (4096 $\rightarrow$ 1000) & 4,097,000 \\
		(40): & SoftMax & \\ \hline
		Total : & \textbf{16 layers} & \textbf{138,357,544} \\
		\hline
	\end{tabular}}
	\caption{Deep architecture used in our experiments. A ReLU non-linearity follows each convolutional and fully-connected layers, beside the last one.
		Convolution (512, 512x3x3, 1x1, 1x1) means 512 filters (e.g. 512 output channels), a kernel size of 256x3x3, 1 step of the convolution to the width and height dimensions, 1 additional zero padded per width to the input, 1 per hight.
		MaxPooling (2,2,2,2) means a 2D pooling operation on 2x2 pixels neighborhood, by step size of 2x2}
	\label{table:Vgg16}
\end{table}

\chapter{InceptionV3}

\begin{table}[h]
	\centering
	\resizebox{290pt}{!}{%
	\begin{tabular}{|c|c|c|}
		\hline
		Layer id & Layer type & Parameters number \\ \hline \hline
		(0): & Image (3, 299, 299) & \\
		(1): & Convolution (32, 3x3x3, 2x2, 0x0) no bias & 864 \\
		(2): & Batch Normalization & 64* \\
		(4): & Convolution (32, 32x3x3, 1x1, 0x0) no bias & 9,216 \\
		(5): & Batch Normalization & 64* \\
		(7): & Convolution (64, 32x3x3, 1x1, 1x1) no bias & 18,432 \\
		(8): & Batch Normalization & 128* \\
		(10): & MaxPooling (3x3,2x2) & \\
		(11): & Convolution (80, 64x3x3, 1x1, 0x0) no bias &5,120 \\
		(12): & Batch Normalization & 160*\\
		(14): & Convolution (192, 80x3x3, 1x1, 0x0) no bias & 138,240 \\
		(15): & Batch Normalization & 384* \\
		(17): & MaxPooling (3x3,2x2) & \\
		(18): & 4 x Inception1 & - \\
		(19): & 4 x Inception2 & - \\
		(20): & Inception3 & - \\
		(20): & 2 x Inception4 & - \\
		(21): & MaxPooling (8x8,1x1) & \\
		(22): & FullyConnected (2048 $\rightarrow$ 1000) & 2,065,392\\
		(23): & SoftMax & \\ \hline
		Total : & \textbf{42 layers} & \textbf{23,816,528} \\
		\hline
	\end{tabular}}
	\caption{Inception architecture used in our experiments. A ReLU non-linearity follows each convolutional and fully-connected layers, beside the last one. * means that the parameters are not learned by backpropagation. For layers Inception1, Inception2, Inception3 and Inception4, please refer to the original paper \cite{szegedy2015rethinking}.
		Convolution (512, 512x3x3, 1x1, 1x1) means 512 filters (e.g. 512 output channels), a kernel size of 256x3x3, 1 step of the convolution to the width and height dimensions, 1 additional zero padded per width to the input, 1 per hight.
		MaxPooling (2,2,2,2) means a 2D pooling operation on 2x2 pixels neighborhood, by step size of 2x2.}
	\label{table:Vgg16}
\end{table}

\end{appendices}

\bibliographystyle{plain}
\bibliography{bibtex}{}

\end{document}